\documentclass{bmvc2k}

\title{Finding Directions in GAN’s Latent Space for Neural Face Reenactment}

\addauthor{Stella Bounareli}{k2033759@kingston.ac.uk}{1}
\addauthor{Vasileios Argyriou}{vasileios.argyriou@kingston.ac.uk}{1}
\addauthor{Georgios Tzimiropoulos}{g.tzimiropoulos@qmul.ac.uk}{2}

\addinstitution{
 Kingston University, London, UK
}
\addinstitution{
 Queen Mary University of London, UK
}

\runninghead{Bounareli et al.}{Finding Directions in GAN’s Latent Space for NFR}

\usepackage{amsmath}
\usepackage{amsfonts}
\usepackage{enumitem}

\begin{document}

\maketitle

\begin{abstract}
This paper is on face/head reenactment where the goal is to transfer the facial pose (3D head orientation and expression) of a target face to a source face. Previous methods focus on learning embedding networks for identity and pose disentanglement which proves to be a rather hard task, degrading the quality of the generated images. We take a different approach, bypassing the training of such networks, by using (fine-tuned) pre-trained GANs which have been shown capable of producing high-quality facial images. Because GANs are characterized by weak controllability, the core of our approach is a method to discover which directions in latent GAN space are responsible for controlling facial pose and expression variations. We present a simple pipeline to learn such directions with the aid of a 3D shape model which, by construction, already captures disentangled directions for facial pose, identity and expression. Moreover, we show that by embedding real images in the GAN latent space, our method can be successfully used for the reenactment of real-world faces. Our method features several favorable properties including using a single source image (one-shot) and enabling cross-person reenactment. Our qualitative and quantitative results show that our approach often produces reenacted faces of significantly higher quality than those produced by state-of-the-art methods for the standard benchmarks of VoxCeleb1 \& 2. Source code is available at: \url{https://github.com/StelaBou/stylegan_directions_face_reenactment}

\end{abstract}

\section{Introduction}

This paper is on face/head reenactment where the goal is to transfer the facial pose, defined here as the rigid 3D face/head orientation \textit{and} the deformable facial expression, of a target facial image to a source facial image. Such technology is the key enabler for creating high-quality digital head avatars which find a multitude of applications in telepresence, Augmented Reality/Virtual Reality (AR/VR), and the creative industries. Recently, and thanks to the advent of Deep Learning, there has been tremendous progress in the so-called neural face reenactment~\cite{burkov2020neural,zakharov2020fast,wang2021one}. Despite the progress, synthesizing photorealistic face/head sequences is still considered a hard task with the quality of existing solutions being far from sufficient for the demanding applications mentioned above.

A major challenge that most prior methods \cite{bao2018towards, zeng2020realistic, zakharov2019few,zakharov2020fast,burkov2020neural,ha2020marionette} have focused so far is how to achieve identity and pose disentanglement to both preserve the appearance and identity characteristics of the source face and successfully transfer to the facial pose of the target face. Training conditional generative models to produce embeddings with such disentanglement properties is known to be a difficult machine learning task \cite{deng2020disentangled,KowalskiECCV2020,shoshan2021gan}, and this turns out to be a significant technical impediment for face reenactment too. Additionally, previous methods~\cite{zakharov2019few, zakharov2020fast} have approached reenactment using paired data during training. However, under such a paired setting it is not clear how to formulate cross-person reenactment \cite{zakharov2019few}. 

In this work, we are taking a different path to neural face reenactment. A major motivation for our work is that unconstrained face generation using modern state-of-the-art GANs~\cite{karras2019style,karras2020analyzing,karras2020training} has reached levels of unprecedented realism to the point that it is often impossible to distinguish real facial images from generated ones. Hence, the research question we would like to address in this paper is: \textit{Can a pretrained GAN}~\cite{karras2020analyzing} \textit{be adapted for face reenactment?} A key challenge that needs to be addressed to this end, is that GANs come with no semantic parameters to control their output. Hence, in order to alleviate this, the core of our approach is a method to discover which \textit{directions in the latent GAN space} are responsible for controlling facial pose and expression variations. Knowledge of these directions would directly equip the pretrained GAN with the desired reenactment capabilities. Inspired by Voynov and Babenko~\cite{voynov2020unsupervised}, we present a very simple pipeline to learn such directions with the help of a linear 3D shape model \cite{feng2020deca}. By construction, such a shape model captures disentangled directions for facial pose, identity and expression which is exactly what is required for reenactment. Moreover, a second key challenge that needs to be addressed is how to use the GAN for the manipulation of real-world images. Capitalizing on \cite{tov2021designing}, we further show that by embedding real images in the GAN latent space, our pipeline can be successfully used for real face reenactment. Overall, we make the following \textbf{contributions}:
\begin{enumerate}[noitemsep]
    \item 
    Instead of training conditional generative models~\cite{burkov2020neural,zakharov2020fast}, we present a different approach to face reenactment by finding the directions in the latent space of a pretrained GAN (StyleGAN2~\cite{karras2020analyzing} fine-tuned on the VoxCeleb1 dataset) that are responsible for controlling the facial pose (i.e. rigid head orientation and expression), and show how these directions can be used for neural face reenactment on video datasets. 
    \item 
    To achieve our goal, we describe \textit{a simple pipeline} that is trained with the aid of a linear 3D shape model which already contains disentangled directions for facial shape in terms of pose, identity and expression. We further show that our pipeline can be trained with real images too by firstly embedding them into the GAN space, enabling the successful reenactment of real-world faces.
    \item
    We show that our method features several favorable properties including using a single source image (one-shot), and enabling cross-person reenactment. 
    \item
    We perform several qualitative and quantitative comparisons with recent state-of-the-art reenactment methods, illustrating that our approach often produces reenacted faces of significantly higher quality for the standard benchmarks of VoxCeleb1 \& 2~\cite{Nagrani17,Chung18b}.
\end{enumerate}


\section{Related work}
\noindent \textbf{Semantic face editing:} There is a plethora of recent works that investigate the existence of interpretable directions in the GAN's latent space \cite{shen2020interfacegan,2020ganspace,voynov2020unsupervised,shen2020closedform, oldfield2021tensor,tzelepis2021warpedganspace,yao2021latent,yang2021discovering, oldfield2022panda,tzelepis2022contraclip}. These methods are able to successfully edit synthetic images, however, most of them do not allow controllable editing and thus they cannot be applied on the face reenactment task. Voynov and Babenko~\cite{voynov2020unsupervised}, introduce an unsupervised method that is able to discover disentangled linear directions in the latent GAN space by jointly learning the directions and a classifier that learns to predict which direction is responsible for the image transformation. Our method is inspired by Voynov and Babenko~\cite{voynov2020unsupervised}, extending it in several ways to make it suitable for neural face reenactment. Additionally, there is a line of work that allows explicit controllable facial image editing ~\cite{deng2020disentangled,gif, durall2021facialgan, shoshan2021gan, wang2021cross, nitzan2020face, abdal2021styleflow}. Related to our method is StyleRig~\cite{tewari2020stylerig} which uses 3DMM parameters to control the generated images from a pretrained StyleGAN2. StyleRig's training pipeline is not end-to-end and significantly more complicated than ours, while in order to learn better disentangled directions, StyleRig requires different models for different attributes (e.g. pose, expression). In contrast, we learn all disentangled directions for face reenactment simultaneously and our model can successfully edit all attributes as well as edit only one attribute. Moreover, StyleRig is mainly applied on synthetic images, thus real image editing is not straightforward. Consequently, the aforementioned issues restrict StyleRig's applicability for real-world face reenactment, where various facial attributes change simultaneously. A follow-up work, PIE~\cite{tewari2020pie}, focuses on inverting real images to enable editing using StyleRig~\cite{tewari2020stylerig}. However, their method is computationally expensive (10 min/image) which is prohibitive for video-based facial reenactment. 

\noindent \textbf{GAN inversion:} GAN inversion aims to encode real images into the latent space of pretrained GANs~\cite{karras2019style,karras2020analyzing}, which enables their editing using existing methods of synthetic image manipulation. Most of the inversion techniques~\cite{alaluf2021restyle,richardson2021encoding,tov2021designing, alaluf2022hyperstyle, dinh2022hyperinverter,wang2022high} train encoder-based architectures that focus on predicting the best latent codes that can generate images visually similar with the original ones and allow successful editing. The authors of~\cite{zhu2020domain} propose a hybrid approach which consists of learning an encoder followed by an optimization step on the latent space to refine the similarity between the reconstructed and real images. Additionally, Richardson \textit{et al.}~\cite{richardson2021encoding} introduce a method that tries to solve the editability-perception trade-off, while recently in~\cite{roich2021pivotal}, the authors propose fine-tuning the generator to better capture appearance features, so that the inverted images resemble the original ones.

\noindent \textbf{Neural face/head reenactment:} Face reenactment is a non-trivial task, as it requires wide generalization across identity and facial pose. Many of the proposed methods rely on facial landmark information \cite{zakharov2019few, tripathy2020icface, zhang2020freenet,ha2020marionette, tripathy2021facegan, zakharov2020fast, wang2022latent, hsu2022dual}. The authors of~\cite{zakharov2020fast} propose a one-shot face reenactment method driven by landmarks, which decomposes an image on pose-dependent and pose-independent components. A limitation of landmark based methods is that landmarks preserve identity information, thus impeding their applicability on cross-subject face reenactment~\cite{burkov2020neural}. In~\cite{burkov2020neural} the authors perform face reenactment by learning pose and identity embeddings using two different encoders. Additionally, warping-based methods~\cite{wiles2018x2face,siarohin2019first,wang2021one,ren2021pirenderer} synthesize the reenacted images based on the motion of the driving faces. Those methods produce realistic results, however they suffer from visual artifacts and pose mismatch especially in large head pose variations. Finally, the authors of~\cite{meshry2021learned} propose a two-step architecture that aims to disentangle the spatial and style components of an image that leads to better preservation of the source identity, while in~\cite{doukas2020headgan} the authors present a GAN-based method conditioned on a 3D face representation~\cite{zhu2016face}.

To summarize, all the aforementioned methods rely on training \textit{conditional} generative models on large paired datasets in order to learn facial descriptors with disentanglement properties. In this paper, we propose a new approach for face reenactment that learns disentangled directions in the latent space of a pretrained StyleGAN2 on the VoxCeleb dataset. We show that the discovery of meaningful and disentangled directions that are responsible for controlling the facial pose can be used for high quality self- and cross-identity reenactment. 

\begin{figure*}[!ht]
\begin{center}
\includegraphics[width=1.0\textwidth]{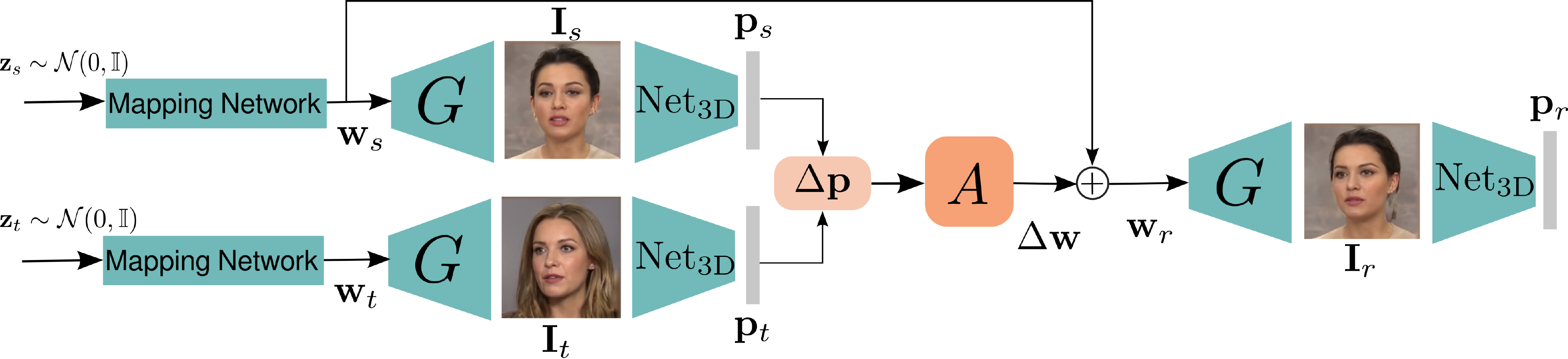}
\end{center}
\caption{\textbf{Overview of our method:} Given a pair of source $\mathbf{I}_s$ and target $\mathbf{I}_t$ images, we calculate the facial pose parameter vectors $\mathbf{p}_s$ and $\mathbf{p}_t$ using the $\mathrm{Net_{3D}}$ network, respectively. The matrix of directions $\mathbf{A}$ is trained such that, given the shift $\Delta \mathbf{w} = \mathbf{A}\Delta \mathbf{p}$, the reenacted image $\mathbf{I}_r$ generated using the latent code $\mathbf{w}_r = \mathbf{w}_s + \boldsymbol{\Delta}\mathbf{w}$, illustrates the facial pose of the target face, while maintaining the identity of the source face.} 
\label{fig:architecture}
\end{figure*}

\section{Method}

Our method consists of three parts detailed in the following subsections. 
In Section~\ref{ssec:latent}, we show how to find the facial pose directions in the latent GAN space and use them for face/head reenactment. In Section~\ref{ssec:real}, we describe how to extend our method to handle real facial images. Finally, in Section~\ref{ssec:video}, we investigate how better results can be obtained by fine-tuning on paired video data.

\subsection{Finding the reenactment latent directions} \label{ssec:latent}

The generator $G$ takes as input latent codes $\mathbf{z} \sim \mathcal{N}(0,\mathbb{I}) \in  \mathbb{R}^{512}$ and generates images $\mathbf{I} = G(\mathbf{z}) \in \mathbb{R}^{3\times256\times256}$. StyleGAN2 firstly maps the latent code $\mathbf{z}$ into the intermediate latent code $\mathbf{w} \in \mathbb{R}^{512}$ using a series of fully connected layers. Then, the latent code $\mathbf{w}$ is fed into each convolution layer of StyleGAN2's generator. This mapping enforces the disentangled representation of StyleGAN2~\cite{karras2020analyzing}. In order to fairly compare our work with previous face reenactment methods, we need a StyleGAN2 model that generates synthetic images that resemble the distribution of the VoxCeleb dataset~\cite{Nagrani17}. This dataset is more diverse compared to Flickr-Faces-HQ (FFHQ) dataset~\cite{karras2019style} in terms of head poses and expressions, providing the ability to find more meaningful directions for face reenactment (e.g. GANs pretrained on FFHQ do not account for roll changes in head pose). Having a pretrained StyleGAN2 generator on FFHQ dataset, we use the method of Karras \textit{et al.}~\cite{karras2020training} to fine-tune the generator on VoxCeleb. We note that we do not finetune the generator under any reenactment objective. Our generator on VoxCeleb is able to produce synthetic images with random identities (different from the identities of VoxCeleb) that follow the distribution of VoxCeleb dataset in terms of head poses and expressions.  

A facial shape $\textbf{s}\in\mathbb{R}^{3N}$ ($N$ is the number of vertices) can be written in terms of a linear 3D shape model as:

\begin{equation}\label{eq:3dmm}
\textbf{s} = \bar{\textbf{s}} + \textbf{S}_{i}\mathbf{p}_{i} +  \textbf{S}_{e}\mathbf{p}_{e},
\end{equation}
where $\bar{\textbf{s}}$ is the mean 3D shape, $\textbf{S}_{i}\in\mathbb{R}^{3N\times m_{i}}$ and $\textbf{S}_{e}\in\mathbb{R}^{3N\times m_{e}}$ are the PCA bases for identity and expression, and $\mathbf{p}_{i}$ and $\mathbf{p}_{e}$ are the corresponding identity and expression coefficients. 
Moreover, we denote as $\mathbf{p}_{\theta} \in \mathbb{R}^{3}$ the rigid head orientation defined by the three Euler angles (yaw, pitch, roll). For reenactment, we are interested in manipulating head orientation and expression, so our facial pose parameter vector is $\mathbf{p}=[\mathbf{p}_{\theta}, \mathbf{p}_{e}] \in \mathbb{R}^{3+m_e}$. We note that all PCA shape bases are orthogonal to each other, and hence they capture disentangled variations of identity and expression. They are calculated in a frontalized reference frame, thus they are also disentangled with head orientation. These bases can be also interpreted as directions in the shape space. We propose to learn similar directions in the GAN latent space. 

In particular, we propose to associate a change $\Delta \mathbf{p}$ in facial pose, with a change $\Delta \mathbf{w}$ in the (intermediate) latent GAN space so that the two generated images $G(\mathbf{w})$ and $G(\mathbf{w}+\Delta \mathbf{w})$ differ only in pose by the same amount $\Delta \mathbf{s}$ induced by $\Delta \mathbf{p}$. If the directions sought in the GAN latent space are assumed to be linear~\cite{nitzan2021large}, this implies the following linear relationship:
\begin{equation}
\Delta \mathbf{w} = \mathbf{A}\Delta \mathbf{p}, \label{Eq:Dw}
\end{equation}
where $\mathbf{A}\in \mathbb{R}^{d_{\mathrm{out}}\times d_{\mathrm{in}}}$ is a matrix, the columns of which (i.e. $d_{\mathrm{in}}$) represent the directions in GAN latent space. In our case, $d_{\mathrm{in}} = (3+m_e)$ and $d_{\mathrm{out}} = N_{l}\times512$, where $N_{l}$ is the number of the generators layers we opt to apply shift changes.

\noindent\textbf{Training pipeline:} The matrix $\mathbf{A}$ is unknown so we propose the simple pipeline of Fig.~\ref{fig:architecture} to estimate it: in particular, we sample two random latent codes $\mathbf{z}_s$ and $\mathbf{z}_t$ ($s$, $t$ for source and target, respectively) and pass them through the generator $G$. The two generated images $\mathbf{I}_s = G(\mathbf{z}_s)$ and $\mathbf{I}_t= G(\mathbf{z}_t)$ are fed into the pre-trained $\mathrm{Net_{3D}}$ which estimates the corresponding pose parameter vectors, $\mathbf{p}_s$ and $\mathbf{p}_t$, respectively. Using Eq.~\ref{Eq:Dw}, we compute $\Delta \mathbf{w} = \mathbf{A}\Delta \mathbf{p}=\mathbf{A}(\mathbf{p}_t-\mathbf{p}_s)$ and $\mathbf{w}_{r} = \mathbf{w}_s + \Delta \mathbf{w}$.
From $\mathbf{w}_{r}$ our pipeline generates the reenacted facial image
$\mathbf{I}_{r}= G(\mathbf{w}_{r})$, which depicts the source face in the facial pose of the target. The only trainable quantity in the above pipeline is the matrix $\mathbf{A}$ containing the unknown directions in GAN latent space. We propose to learn it in a self-supervised manner. 

\noindent\textbf{Training losses:} Our pipeline is trained by minimizing the following total loss:
\begin{equation}\label{eq:loss_all}
\mathcal{L} = \lambda_{r} \mathcal{L}_{r} + \lambda_{id} \mathcal{L}_{id} + \lambda_{per} \mathcal{L}_{per},
\end{equation}
where $\lambda_{r} = 1,  \lambda_{id} = 10$ and $\lambda_{per} = 10$. The \textit{reenactment loss} $\mathcal{L}_{r}$ ensures successful facial pose transfer from target to source and it is defined as: $\mathcal{L}_{r} =  \mathcal{L}_{sh} + \mathcal{L}_{eye} + \mathcal{L}_{mouth}$. $\mathcal{L}_{sh} = \|\mathbf{S}_r - \mathbf{S}_{gt}\|_1$ is the shape loss, 
where $\mathbf{S}_r$ is the 3D shape of the reenacted image and $\mathbf{S}_{gt}$ is the reconstructed \textit{ground-truth} 3D shape calculated using Eq.~\ref{eq:3dmm} with the identity coefficients $\mathbf{p}_{i}$ of the source face and the coefficients $\mathbf{p}_{e}$ of the target face. Additionally, to enhance the expression transfer we calculate the eye loss $\mathcal{L}_{eye}$ (the mouth loss $\mathcal{L}_{mouth}$ is computed in a similar fashion) which compares the inner distances between the eye landmark pairs of upper and lower eyelids between the reenacted and reconstructed ground-truth shapes (see supplementary for detailed explanation of eye $\mathcal{L}_{eye}$ and mouth $\mathcal{L}_{mouth}$ losses). Additionally, $\mathcal{L}_{id}$ is an \textit{identity loss} based on the cosine similarity of features extracted from the source $\mathbf{I}_s$ and the reenacted $\mathbf{I}_r$ image using the face recognition network of~\cite{deng2019arcface}. This loss imposes that the identity of the source is preserved in the reenacted image. Finally, we also found that better image quality is obtained if we additionally use $\mathcal{L}_{per}$ which is the standard \textit{perceptual loss} of \cite{johnson2016perceptual}. 

\noindent\textbf{Training details:} We estimate the distribution of each element of the pose parameters $\mathbf{p}$ by randomly generating $10\mathrm{K}$ images and computing their corresponding $\mathbf{p}$ vectors. Using the estimated distributions, during training, we re-scale each element of $\mathbf{p}$ from its original range to a common range $[-a,a]$. Furthermore, to increase the disentanglement of the learned directions of our method, we follow a training strategy where for $50\%$ of the training samples we reenact only one attribute by using $\Delta \mathbf{p} = [0,...,\varepsilon_i,...0]$, where $\varepsilon_i$ is sampled from a uniform distribution $\mathcal{U}[-a, a]$.

\subsection{Real image reenactment}
\label{ssec:real}

So far our method is able to transfer facial pose from a source facial image to a target only for synthetically generated images. To extend our method to work with real images, in this section, we propose (a) to use a pipeline for inverting the images back to the latent code space of StyleGAN2, and (b) a mixed training approach for discovering the latent directions.

\noindent \textbf{Real image inversion:} Ideally, the inversion method should produce latent codes that can generate facial images identical with the original ones \textit{and} enable image editing without producing visual artifacts. Although satisfying both requirements is challenging~\cite{alaluf2021restyle,richardson2021encoding,tov2021designing}, we found that the following pipeline produces excellent results for the purposes of our goal (i.e. face/head reenactment). During training, we employ an encoder based method~\cite{tov2021designing} to invert the real images into the $\mathcal{W}+$ space~\cite{abdal2019image2stylegan}. However, directly using the inverted latent codes $\mathbf{w}^{inv}$ does not produce satisfactory reenactment results. This happens because the latent codes obtained from inversion, may present a domain gap from the latent codes of synthetic images. To alleviate this, we propose a mixed data approach for training the pipeline of Section~\ref{ssec:latent}: specifically, we first invert the extracted frames from the VoxCeleb dataset, and during training, at each iteration (i.e. for each batch) we use 50\% random latent codes $\mathbf{w}$ and 50\% embedded latent codes $\mathbf{w}^{inv}$. 

The inverted images using the encoder based method~\cite{tov2021designing} might still be missing some identity details. To alleviate this, only during inference, we use an additional optimization step~\cite{roich2021pivotal} that lightly optimizes the generator, so that the newly generated image more closely resembles the original one. Note that this step does not affect the calculation of $\mathbf{w}^{inv}$ and is used only during inference to obtain a higher quality inversion. We perform the optimization for 200 steps and only on the source frame of each video.

\subsection{Fine-tuning on paired video data}
\label{ssec:video}

Our method so far has been trained with unpaired static facial images. This has at least two advantages: (a) it enables training with a very large number of identities, and (b) seems more suitable for cross-person reenactment. However, additional improvements enabled by the optimization of additional losses can be obtained by further training on paired data from VoxCeleb1. Compared to training from scratch on video data, as in most previous methods (e.g.~\cite{zakharov2019few,zakharov2020fast,burkov2020neural}), we believe that our approach offers a more balanced option that combines the best of both worlds: training with unpaired static images and fine-tuning with paired video data. From each video of our training set, we randomly sample a source and a target face that have the same identity but different pose and expression. Consequently, we minimize the following loss function $\mathcal{L} = \lambda_{r}\mathcal{L}_{r} + \lambda_{id}\mathcal{L}_{id} + \lambda_{per}\mathcal{L}_{per} + \lambda_{pix}\mathcal{L}_{pix}$,
where $\mathcal{L}_{r}$ is the same reenactment loss defined in Section~\ref{ssec:latent}, $\mathcal{L}_{id}$ and $\mathcal{L}_{per}$ are the identity and perceptual losses, however this time calculated between the reenacted $\mathbf{I}_r$ and the target image $\mathbf{I}_t$ and $\mathcal{L}_{pix}$ is a pixel-wise $L1$ loss between the reenacted and target images. 

\section{Experiments}
\label{sec:exp}

In this section, we present qualitative and quantitative results and comparisons of our method with recent state-of-the-art approaches. The bulk of our results and comparisons, reported in Section~\ref{ssec:face_re}, are on self- and cross-person reenactment on the VoxCeleb1~\cite{Nagrani17} dataset. Comparisons with state-of-the-art on the VoxCeleb2~\cite{Chung18b} test set released by~\cite{zakharov2019few} are provided in the supplementary material. Finally, in Section~\ref{ssec:ablation} we report ablation studies on the various design choices of our method. 

\noindent \textbf{Implementation details:} We fine-tune StyleGAN2 on the VoxCeleb1 dataset with $256\times256$ image resolution and we train the encoder of~\cite{tov2021designing} for real image inversion. The 3D shape model we use is DECA~\cite{feng2020deca}. For our training procedure, we only learn a matrix of directions $\mathbf{A}\in \mathbb{R}^{(N_l\times512) \times k}$ where $k=3+m_e, m_e = 12$ and $N_l=8$. We train three matrices of directions: the first one is on synthetically generated images (Section~\ref{ssec:latent}), while the second one on mixed real and synthetic data (Section~\ref{ssec:real}). Finally, as described in Section~\ref{ssec:video}, we obtain a third model by fine-tuning the second one on paired data. For training, we used the Adam optimizer~\cite{kingma2014adam} with constant learning rate $10^-4$. We train our models for 20K iterations with a batch size of 12 on synthetic and real images. Fine-tuning is performed on real paired images for 150K iterations. All models are implemented in PyTorch~\cite{paszke2017automatic}.

\subsection{Comparison with state-of-the-art on VoxCeleb}
\label{ssec:face_re}

Herein, we compare the performance of our method against the state-of-the-art in face reenactment on VoxCeleb1~\cite{Nagrani17}. We conduct two types of experiments, namely self- and cross-person reenactment. For evaluation purposes, we use both the video data provided by~\cite{zakharov2019few} and the original test-set of VoxCeleb1. We note that there is no overlap between the train and test identities and videos. Similar comparisons on the VoxCeleb2~\cite{Chung18b} test set released by~\cite{zakharov2019few} are provided in the supplementary material. We compare our method quantitatively and qualitatively with six methods: X2Face~\cite{wiles2018x2face}, FOMM~\cite{siarohin2019first}, Fast bi-layer~\cite{zakharov2020fast}, Neural-Head~\cite{burkov2020neural}, LSR~\cite{meshry2021learned} and PIR~\cite{ren2021pirenderer}.
For X2Face~\cite{wiles2018x2face}, FOMM~\cite{siarohin2019first} and PIR~\cite{ren2021pirenderer}, we use the pretrained (by the authors) model on VoxCeleb1. For Fast bi-layer~\cite{zakharov2020fast}, Neural-Head~\cite{burkov2020neural} and LSR~\cite{meshry2021learned} we also use the pretrained (by the authors) models on VoxCeleb2~\cite{Chung18b}. For fair comparison with the methods of Neural-Head~\cite{burkov2020neural} and LSR~\cite{meshry2021learned}, we evaluate their model under the one-shot setting.

\noindent \textbf{Quantitative comparisons:} We report seven different metrics. We compute the Learned Perceptual Image Path Similarity (LPIPS)~\cite{zhang2018unreasonable} to measure the perceptual similarities, and to quantify identity preservation we compute the cosine similarity (CSIM) of ArcFace~\cite{deng2019arcface} features. Moreover, we measure the quality of the reenacted images using the Frechet-Inception Distance (FID) metric~\cite{heusel2017gans}, while we also report the Fréchet Video Distance (FVD)~\cite{unterthiner2018towards} metric that measures both the video quality and the temporal consistency of the generated videos. To quantify the facial pose transfer, we calculate the normalized mean error (NME) between the predicted landmarks in the reenacted and target images. We use~\cite{bulat2017far} for landmark estimation, and we calculate the NME by normalizing it with the square root of the ground truth face bounding box and multiplying it by $10^3$. We further evaluate pose transfer by calculating the mean $L1$ distance of the head pose (Pose) in degrees and the mean $L1$ distance of the expression coefficients $\mathbf{p}_{e}$ (Exp.). 

\begin{figure}[h!]
\begin{center}
{\includegraphics[width=0.8\textwidth]{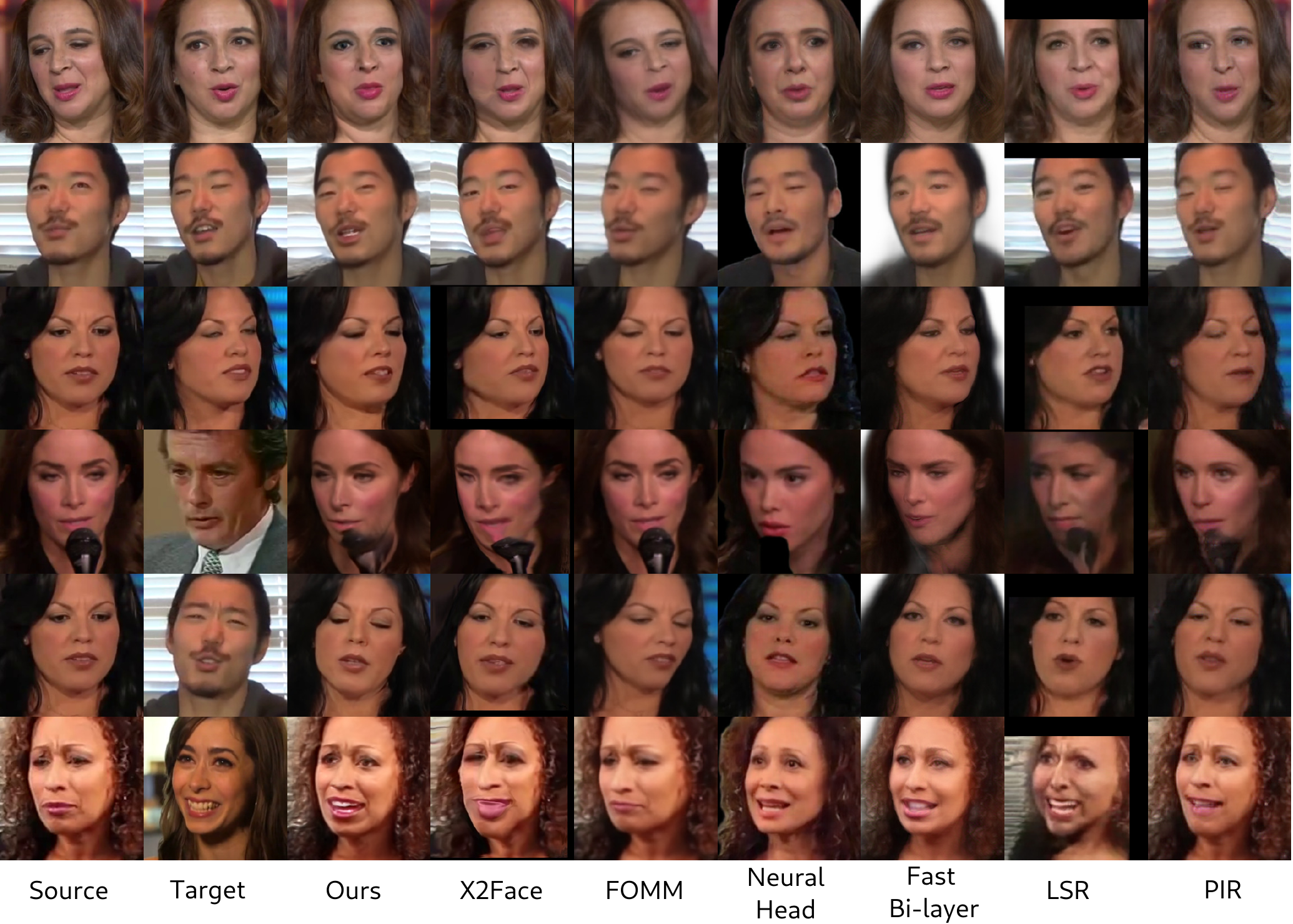}}
\end{center}
  \caption{Qualitative results and comparisons for self (top three rows) and cross-subject reenactment (last three rows) on VoxCeleb1. The first and second columns show the source and target faces. Our method preserves the appearance and identity characteristics (e.g. face shape) of the source face significantly better and also better captures fine-grained expression details such as closed eyes (2\textsuperscript{nd} and 5\textsuperscript{th} row).}
\label{fig:reenactment}
\end{figure}

In Tables~\ref{table:self_reenactment_metrics} and~\ref{table:cross_reenactment_metrics}, we report the quantitative results for self and cross-subject reenactment, respectively. For self-reenactment, we combine the original test set of VoxCeleb1~\cite{Nagrani17} and the test set provided by~\cite{zakharov2019few}. For cross-subject reenactment, we randomly select 200 video pairs from the small test set of~\cite{zakharov2019few}. In self-reenactment, all metrics are calculated between the reenacted and the target faces, while in cross-subject reenactment, CSIM is calculated between the source and the reenacted faces and pose/expression error between the target and the reenacted faces. As a result, the values of CSIM in cross-subject reenactment are expected to be lower. Regarding self-reenactment, X2Face and PIR have a higher value on CSIM, however we argue that this is due to their warping-based technique which enables better reconstruction of the background and other identity characteristics. Importantly, this quantitative result is accompanied by poor qualitative results (e.g. see Fig.~\ref{fig:reenactment}). Additionally, regarding pose transfer, we achieve similar results on NME and Pose error with Fast Bi-layer~\cite{zakharov2020fast} and LSR~\cite{meshry2021learned} (their methods are trained on VoxCeleb2 which contains 5$\times$ more identities) and we outperform all methods on expression transfer. Finally, our results on FID and FVD metric confirm that the quality of our generated videos resembles the quality of VoxCeleb dataset. Cross-subject reenactment is more challenging, as source and target faces have different identities. In this case, it is important to maintain the source identity characteristics without transferring the target ones. In Table~\ref{table:cross_reenactment_metrics}, the high CSIM value for FOMM is not accompanied by good qualitative results as shown in Fig.~\ref{fig:reenactment}, where FOMM, in most cases, is not able to transfer the target head pose (hence their method achieves higher CSIM but poor pose transfer). Additionally, we achieve better head pose and expression transfer compared to all other methods.

\begin{table}[h]
\begin{center}
\begin{tabular}{|c||c|c|c|c|c|c|c|}
    \hline
    Method & CSIM  & LPIPS &  FID & FVD & NME & Pose  & Exp. \\
    \hline
    X2Face~\cite{wiles2018x2face} & \underline{0.70} & 0.13  & \underline{35.5} & 409 & 17.8 & 1.5 & 0.90 \\
    FOMM~\cite{siarohin2019first} & 0.65 & 0.14 & 35.6 & \underline{402} & 34.1 & 5.0 & 1.3 \\
    Fast Bi-layer~\cite{zakharov2020fast} & 0.64 & 0.23  & 52.8 & 634 & \textbf{13.2} & \underline{1.1} & 0.80 \\
    Neural-Head~\cite{burkov2020neural} & 0.40 & 0.22  & 98.4 & 587 & 15.5 & 1.3 & 0.90 \\
    LSR~\cite{meshry2021learned} & 0.59 & 0.13 & 45.7 & 464 & 17.8 & \textbf{1.0} & \underline{0.75} \\
    PIR~\cite{ren2021pirenderer} & \textbf{0.71} & \underline{0.12} & 57.2 & 414 & 18.2 & 1.86 & 0.94 \\
    Ours & 0.66 & \textbf{0.11}  & \textbf{35.0} & \textbf{345} & \underline{14.1} & \underline{1.1} & \textbf{0.68} \\
    \hline
\end{tabular}
\end{center}
\caption{Quantitative results on self-reenactment. The results are reported on the combined original test set of VoxCeleb1~\cite{Nagrani17} and the test set released by~\cite{zakharov2019few}. For CSIM metric, higher is better ($\uparrow$), while in all other metrics lower is better ($\downarrow$).}\label{table:self_reenactment_metrics}
\end{table}

\begin{table}[h]
\begin{center}
\begin{tabular}{|c||c|c|c|}
    \hline
    Method & CSIM  & Pose  & Exp. \\
    \hline
    X2Face~\cite{wiles2018x2face}  & 0.57 & 2.2 & 1.5\\
    FOMM~\cite{siarohin2019first} & \textbf{0.73} & 7.7 & 2.0\\
    Fast Bi-layer~\cite{zakharov2020fast} & 0.48 & 1.5 & 1.3\\
    Neural-Head~\cite{burkov2020neural}  & 0.36 & 1.7 & 1.6\\
    LSR~\cite{meshry2021learned} & 0.50 & \underline{1.4} & \underline{1.2} \\
    PIR~\cite{ren2021pirenderer} & 0.62 & 2.2 & 1.4  \\
    Ours  & \underline{0.63} &  \textbf{1.2} & \textbf{1.0} \\
    \hline
\end{tabular}
\end{center}
\caption{Quantitative results on cross-subject reenactment. The results are reported on 200 video pairs from the test set of~\cite{zakharov2019few}. For CSIM metric, higher is better ($\uparrow$), while in all other metrics lower is better ($\downarrow$).}\label{table:cross_reenactment_metrics}
\end{table}

\noindent\textbf{Qualitative comparisons:} Unfortunately, quantitative comparisons alone are insufficient to capture the quality of reenactment. Hence, we opt for qualitative visual comparisons \textit{in multiple ways}: (a) results in Fig.~\ref{fig:reenactment}, (b) in the supplementary material, we provide more results in self and cross-subject reenactment both on VoxCeleb1 and VoxCeleb2 datasets, and (c) we also provide \textit{all} self-reenactment videos for the small test set of VoxCeleb1 (and VoxCeleb2) provided in~\cite{zakharov2019few} and cross-reenactment videos of \textit{randomly selected} identities (providing all possible pairs is not possible). As we can see from Fig.~\ref{fig:reenactment} and the videos provided in the supplementary material, our method provides, for the majority of videos, the highest reenactment quality including accurate transfer of pose and expression and, significantly enhanced identity preservation compared to all other methods. Importantly one great advantage of our method on cross-subject reenactment, as shown in Fig.~\ref{fig:reenactment}, is that it is able to reenact the source face with minimal identity leakage (e.g facial shape) from the target face, in contrast to landmark-based methods such as Fast Bi-layer~\cite{zakharov2020fast}. Finally, to show that our method is able to generalise well on other facial video datasets, we provide additional results on the FaceForensics~\cite{roessler2018faceforensics} and 300-VW~\cite{shen2015first} datasets in the supplementary material.

\subsection{Ablation studies}\label{ssec:ablation}

We perform several ablation tests to (a) measure the impact of the identity and perceptual losses, and the additional shape losses for the eyes and mouth, (b) validate our trained models on synthetic, mixed and paired images, and (c) assess the use of optimization in $G$ during inference. For (a), we perform experiments on synthetic images with and without the identity and perceptual losses. To evaluate the models, we randomly generate $5K$ pairs of synthetic images (source and target) and reenact the source image with the pose and expression of the target. As shown in Table~\ref{table:ablation_id}, the incorporation of the identity and perceptual losses is crucial to isolate the latent space directions that strictly control the head pose and expression characteristics without affecting the identity of the source face. In a similar fashion, in Table~\ref{table:ablation_id}, we show the impact of the additional shape losses, namely the eye $\mathcal{L}_{eye}$ and mouth $\mathcal{L}_{mouth}$ losses. As shown, omitting these losses leads to higher pose and expression error.

\begin{table}
\begin{center}
\begin{tabular}{|c||c|c|c|}
\hline
Method & CSIM $\uparrow$ & Pose $\downarrow$ & Exp. $\downarrow$ \\
\hline
Ours w/ $\mathcal{L}_{id} +\mathcal{L}_{per}$  & \textbf{0.52} & 2.4  & 1.2  \\
Ours w/o $\mathcal{L}_{id} + \mathcal{L}_{per}$ & 0.42 & 2.5 & 1.2\\
\hline
Ours w/ $\mathcal{L}_{eye} +\mathcal{L}_{mouth}$ & 0.52 & \textbf{2.4} & \textbf{1.2} \\
Ours w/o $\mathcal{L}_{eye} + \mathcal{L}_{mouth}$  & 0.52 & 2.6 & 1.5\\
\hline
\end{tabular}
\end{center}
\caption{Ablation study on the impact of the identity $\mathcal{L}_{id}$ and perceptual $\mathcal{L}_{per}$ losses, and on the impact of eye $\mathcal{L}_{eye}$ and mouth $\mathcal{L}_{mouth}$ losses. CSIM is calculated between the source and the reenacted images which are on different poses.}
\label{table:ablation_id}
\end{table}

For (b), we evaluate the three different training schemes, namely synthetic only (Section 3.1), mixed synthetic-real (Section 3.2), and mixed synthetic-real fine-tuned with paired data (Section 3.3) for self-reenactment. The results, shown in Table~\ref{table:diff_models} (first three rows), illustrate the impact of each of these training schemes with the one using paired data providing the best results as expected. Finally, regarding (c), we report results of self-reenactment, without any optimization and with optimization of $G$. As shown in Table~\ref{table:diff_models} (last two rows), the optimization of $G$ improves our results (as expected) especially regarding the identity preservation (CSIM). Moreover, with this ablation study we show that our learned directions do not get adversely affected by the optimization step, as both Pose and Expression errors are improving. We note that, to evaluate the different models in Table~\ref{table:diff_models}, we use the small test set of~\cite{zakharov2019few}.

\begin{table}
\begin{center}
\begin{tabular}{|c||c|c|c|}
\hline
Method & CSIM $\uparrow$ & Pose $\downarrow$  & Exp. $\downarrow$ \\
\hline
Ours \textit{synthetic}  & 0.60 & 1.7 & 1.1\\
Ours \textit{real} \& \textit{synthetic} & 0.63 & 1.6 & 1.1\\
Ours \textit{paired} & \textbf{0.66} & \textbf{1.1} & \textbf{0.8}\\
\hline
\textit{w/o optim.}  & 0.45 & 1.4 &  1.0  \\
\textit{w/ optim. in $G$} & \textbf{0.66} & \textbf{1.1} & \textbf{0.8}\\
\hline
\end{tabular}
\end{center}
\caption{Ablation studies on self-reenactment using three different models: (a) trained on synthetic images, (b) trained on both synthetic and real images, and (c) fine-tuned on paired data, and on self reenactment with and without optimization of the generator $G$.}
\label{table:diff_models}
\end{table}

\section{Discussion and conclusions}

This paper introduces a new approach to neural head/face reenactment using a 3D shape model to learn disentangled directions of facial pose in latent GAN space. The approach comes with specific advantages such as the use of powerful pre-trained GANs and 3D shape models which have been studied and developed for several years in computer vision and machine learning. These advantages however, in some cases, can turn into disadvantages. For example, we observed that in extreme source and target poses the reenacted images have some visual artifacts. We attribute this to the GAN inversion process, which renders the inverted latent codes in extreme head poses less editable. Finally, we acknowledge that although face reenactment can be used in a variety of applications such as art, video conferencing etc., it can also be applied for malicious purposes. However, our work does not amplify any of the potential dangers that already exist.


\bibliography{egbib}


\newpage
\section*{Appendix}
\vspace*{-0.075in}
\appendix

\section{Supplementary Material}

In the supplementary material, we first provide an analysis of the discovered directions in the latent space (Section~\ref{ssec:analysis}). Moreover, in Section~\ref{ssec:shape_losses} we describe in detail the computation of the shape losses and in Section~\ref{ssec:limitations} we discuss the limitations of our method. Additionally, we show results of our method on facial attribute editing (Section~\ref{ssec:image_editing}). Finally, in Section~\ref{ssec:results} we provide additional quantitative and qualitative results both on VoxCeleb1~\cite{Nagrani17} and VoxCeleb2~\cite{Chung18b} datasets, we show results on FaceForensics~\cite{roessler2018faceforensics} and 300-VW~\cite{shen2015first} datasets and we compare with state-of-the-art methods for synthetic image editing on FFHQ dataset~\cite{karras2019style}. 

\subsection{Analysis of the learned directions}\label{ssec:analysis}

\textbf{Linearity:} In our work, we discover the disentangled directions that control the facial pose by optimising a matrix $\mathbf{A}$ so that:
\begin{equation}\label{Eq:Dw_s}
    \Delta\mathbf{w}=\mathbf{A}\Delta\mathbf{p}, 
\end{equation}
where $\Delta\mathbf{w}$ denotes a shift in the latent space and $\Delta\mathbf{p}$ denotes the corresponding change in the parameters space. That is, independently of the source attributes, we assume linearity between a shift $\Delta\mathbf{w}$ that is applied to an arbitrary code $\mathbf{w}$ and the induced change $\Delta\mathbf{p}$ in the parameter space -- i.e., the change between the source and the reenacted attributes.

\begin{figure*}
    \begin{center}
    \includegraphics[width=0.8\textwidth]{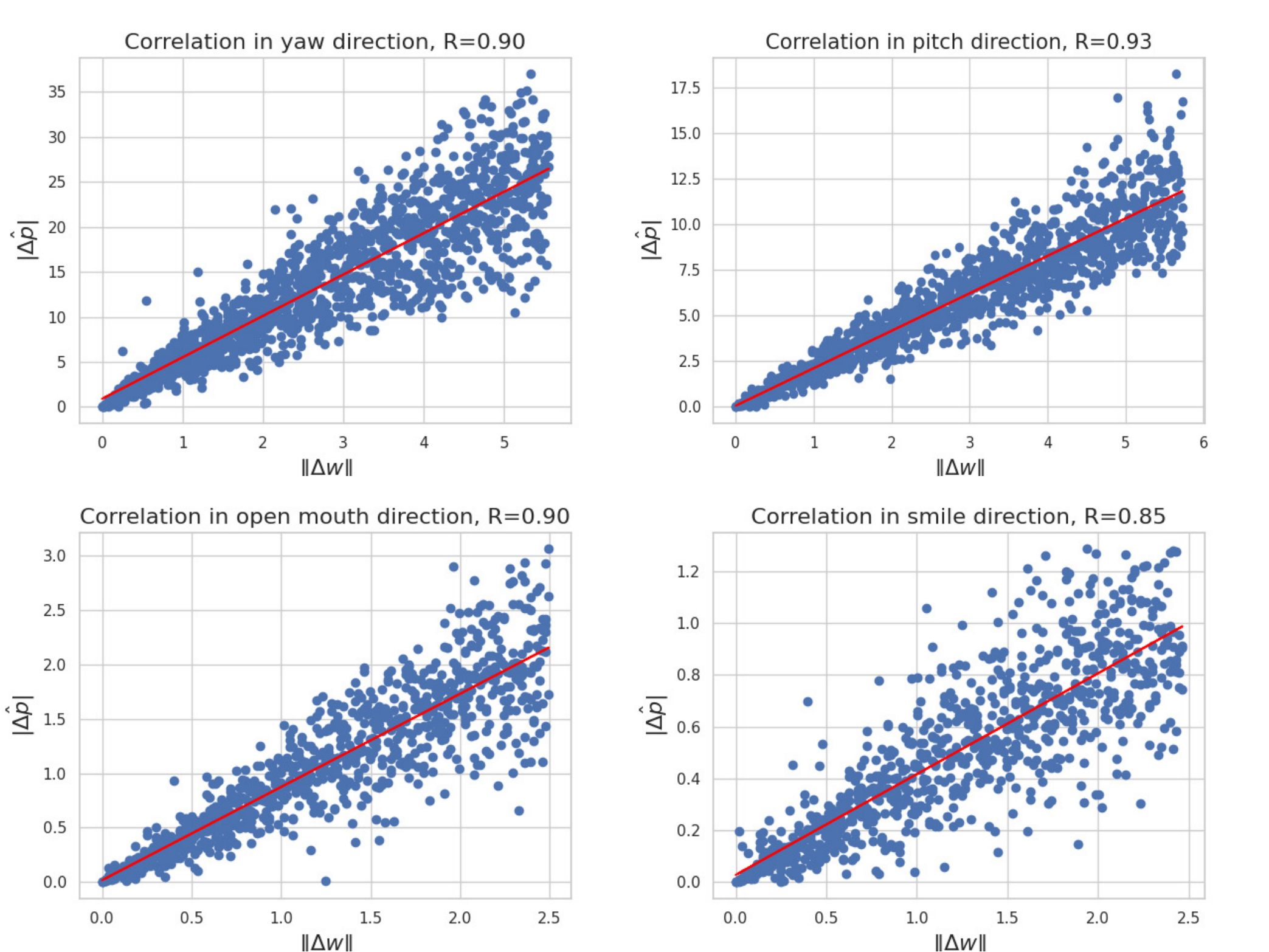}
    \end{center}
      \caption{Analysis of the correlation between shifts $\Vert\Delta\mathbf{w}\Vert$ in the latent space and the predicted changes $\lvert\hat{\Delta\mathbf{p}}\rvert$ in the parameters space. We show results of four different attributes (yaw and pitch angles, smile, and open mouth). In all attributes the correlation is high, indicating strong linear relationship.}  
    \label{fig:linearity}
\end{figure*}

Several recent methods propose to learn linear directions in the latent space of StyleGAN~\cite{voynov2020unsupervised,shen2020interfacegan,shen2020closedform} in order to perform synthetic image editing, based on the fact that the $\mathcal{W}$ latent space of StyleGAN~\cite{karras2019style} has been designed to be linear and disentangled. Furthermore, in~\cite{nitzan2021large}, Nitzan et al. provide a comprehensive analysis on the existence of linear relations between the magnitude of change in the semantic attributes (e.g., pose, smile, etc) and the traversal distance along the corresponding \textit{linear} latent paths. In order to further support our hypothesis (i.e., Eq.~\ref{Eq:Dw_s}), we perform a similar analysis by examining the correlation between random shifts in the latent space, $\Delta\mathbf{w}$, and the predicted shifts in the parameters space, $\hat{\Delta \mathbf{p}}$. Specifically, given a \textit{known} change $\Delta\mathbf{p}$, we calculate the corresponding $\Delta\mathbf{w}$ using Eq.~\ref{Eq:Dw_s} and we apply this change (i.e., $\Delta\mathbf{w}$) onto random latent codes of images with different attributes. Then, we calculate the \textit{predicted} change $\hat{\Delta\mathbf{p}}$ between the source and the reenacted images. In Fig.~\ref{fig:linearity} we demonstrate the results of our analysis in four different attributes, namely, yaw angle, pitch angle, smile, and open mouth. In all attributes, the calculated correlation is close to 0.9 indicating strong linear relationship.

\begin{figure*}[ht]
\begin{subfigure}[b]{1.0\textwidth}
  \centering
  \includegraphics[width=1.0\linewidth]{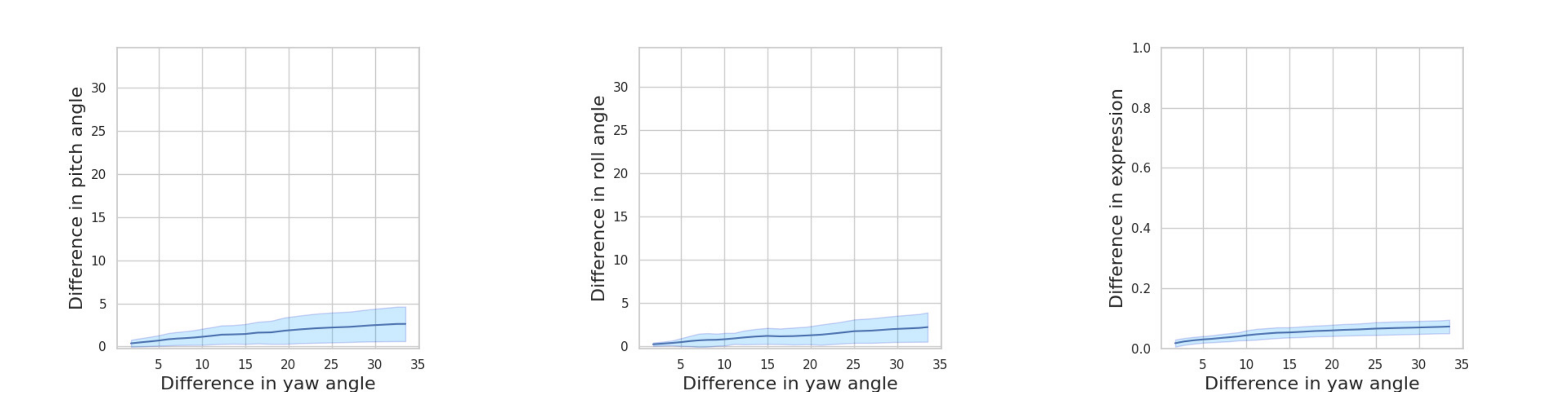}  
  \caption{L1 distance in pitch, roll angles (in degrees) and expression ($\mathbf{p}_{e}$ coefficients) when transferring only the yaw angle from the target images.}
  \label{fig:dis_yaw}
\end{subfigure}
\begin{subfigure}[b]{1.0\textwidth}
  \centering
  \includegraphics[width=1.0\textwidth]{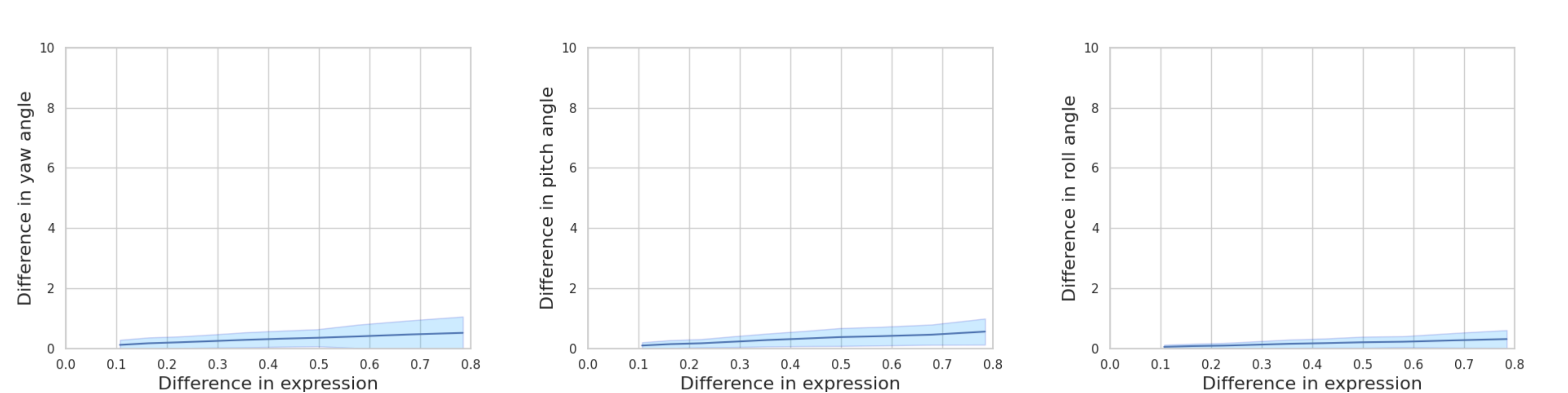}  
  \caption{L1 distance in yaw, pitch and roll angles (in degrees) when transferring only the smile expression from the target images.}
  \label{fig:dis_smile}
\end{subfigure}
\caption{Difference between the source and reenacted facial attributes when transferring only one facial attribute (e.g., yaw angle and smile expression) from the target images.}
\label{fig:disentanglement}
\end{figure*}

\textbf{Disentanglement:} Following the common understanding of disentanglement in the area of GANs~\cite{chen2016infogan,deng2020disentangled,karras2019style}, we refer to a disentangled latent direction, when travelling across it, leads to image generations where only a single attribute changes. To assess the directions learnt by our method in terms of disentanglement, in Fig.~\ref{fig:disentanglement} we illustrate the differences between the source and reenacted attributes when changing a single attribute. In Fig.~\ref{fig:dis_yaw}, we only transfer the yaw angle from the target image, while in Fig.~\ref{fig:dis_smile} we only transfer the smile expression from the target image. We observe that the differences between the rest of the attributes (i.e., pitch, roll, and expression in Fig.~\ref{fig:dis_yaw} and yaw, pitch, and roll in Fig.~\ref{fig:dis_smile}) are clearly small, which indicates that the discovered directions are disentangled. We note that these plots were calculated using 2000 random image pairs. In Fig.~\ref{fig:dis_yaw}, we show the differences in yaw angle that were calculated as the absolute difference between the source and the target yaw angles (measured in degrees), while the differences in the \textit{unchanged} attributes were calculated between the source and reenacted images. In a similar way, in Fig.~\ref{fig:dis_smile} we show the differences in expression that were calculated as the absolute difference between the source and the target expression ($\mathbf{p}_e$ coefficients).

\subsection{Shape losses}\label{ssec:shape_losses}

\begin{figure}
\begin{center}
{\includegraphics[width=0.5\textwidth]{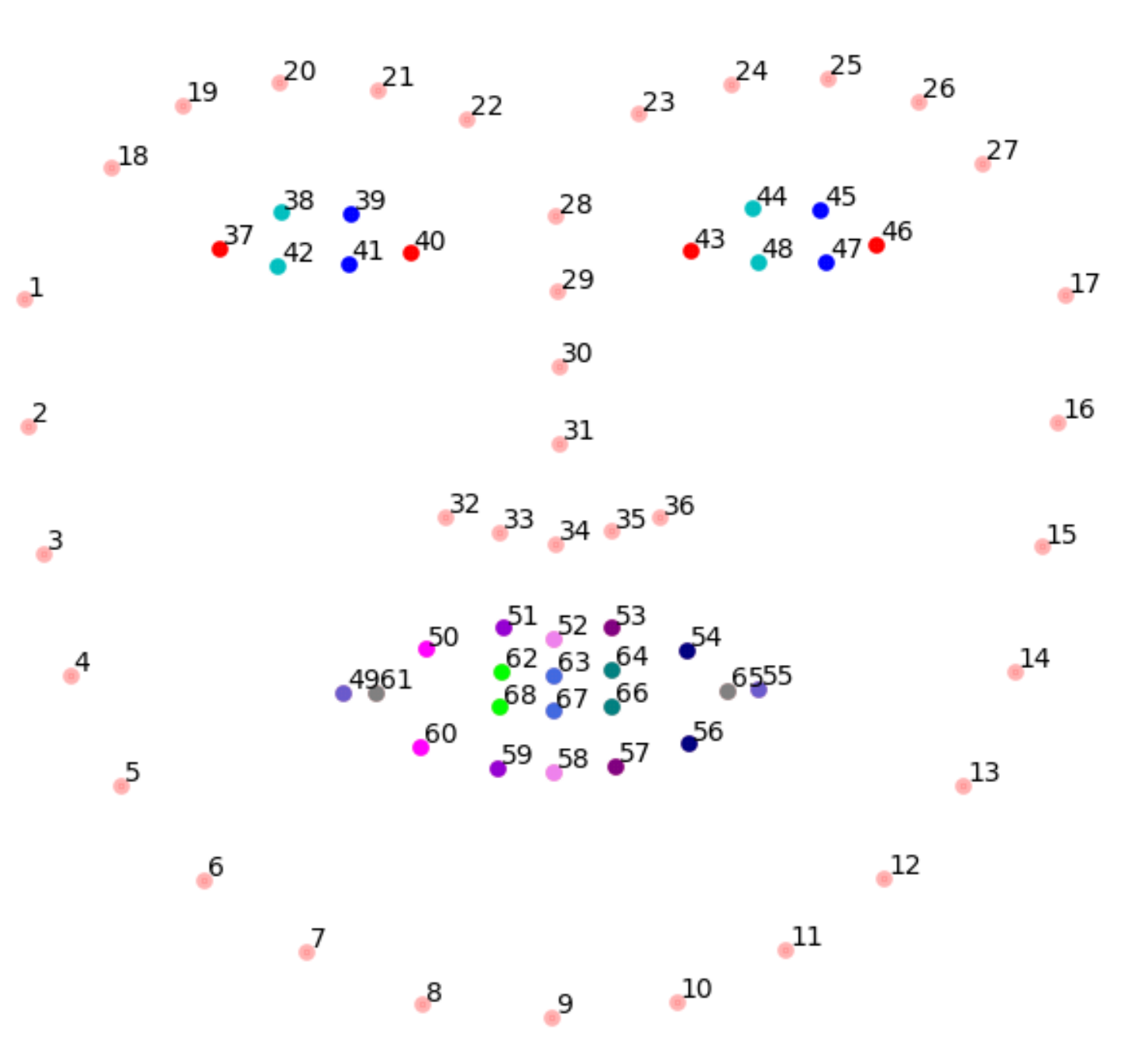}}
  \caption{Depiction of the landmark pairs $P_{eye}$ and $P_{mouth}$ that contribute to the corresponding losses $\mathcal{L}_{eye}$ and $\mathcal{L}_{mouth}$. The landmarks of each pair are drawn with the same color.} 
\label{fig:landmarks_pairs}
\end{center}
\end{figure}

In order to transfer the target facial pose to the source face, we calculate the \textit{reenactment loss} as:
\begin{equation}\label{eq:loss_cat}
\mathcal{L}_{r} =  \mathcal{L}_{sh} + \mathcal{L}_{eye} + \mathcal{L}_{mouth}, 
\end{equation}
where $\mathcal{L}_{sh} $ is the shape loss and $\mathcal{L}_{eye}$, $\mathcal{L}_{mouth}$ the eye and mouth loss, respectively. As shown in our ablation studies (Section 4.2), $\mathcal{L}_{eye}$, $\mathcal{L}_{mouth}$ losses contribute to more accurate expression transfer from the target face to the source face. Specifically, eye loss compares the inner distances $\mathrm{d} = \|(\cdot, \cdot)\|_1$ of the eye landmark pairs (defined as $P_{eye}$) of upper and lower eyelids between the reenacted ($\mathbf{S}_r$) and reconstructed ground-truth ($\mathbf{S}_{gt}$) shapes:
\begin{equation}\label{eq:eye_loss}
\mathcal{L}_{eye} = \sum_{(i,j)\in P_{eye}} \left \| \mathrm{d}\big(\mathbf{S}_r(i),\mathbf{S}_r(j)\big) - \mathrm{d}\big(\mathbf{S}_{gt}(i),\mathbf{S}_{gt}(j) \right \|,
\end{equation}
Similarly, mouth loss is computed between the mouth landmark pairs. In Fig.~\ref{fig:landmarks_pairs}, we show the landmark pairs that are used to calculate these losses. In more detail, $P_{eye}$ and $P_{mouth}$ are defined as:
\begin{equation*}
\begin{split}
P_{eye} = \big[(37,40), (38,42), (39, 41),  \\
 (43,46), (44,48), (45,47) \big],
\end{split}
\end{equation*} 
\begin{equation*}
\begin{split}
P_{mouth} = \big[(49,55), (50,60), (51,59), (52,58),  \\
 (53,57), (54,56), \\ 
 (61,65), (62,68), (63,67), (64,66) \big]
\end{split}
\end{equation*} 

\subsection{Limitations}\label{ssec:limitations}

\begin{figure*}
\begin{center}
\includegraphics[width=1.0\textwidth]{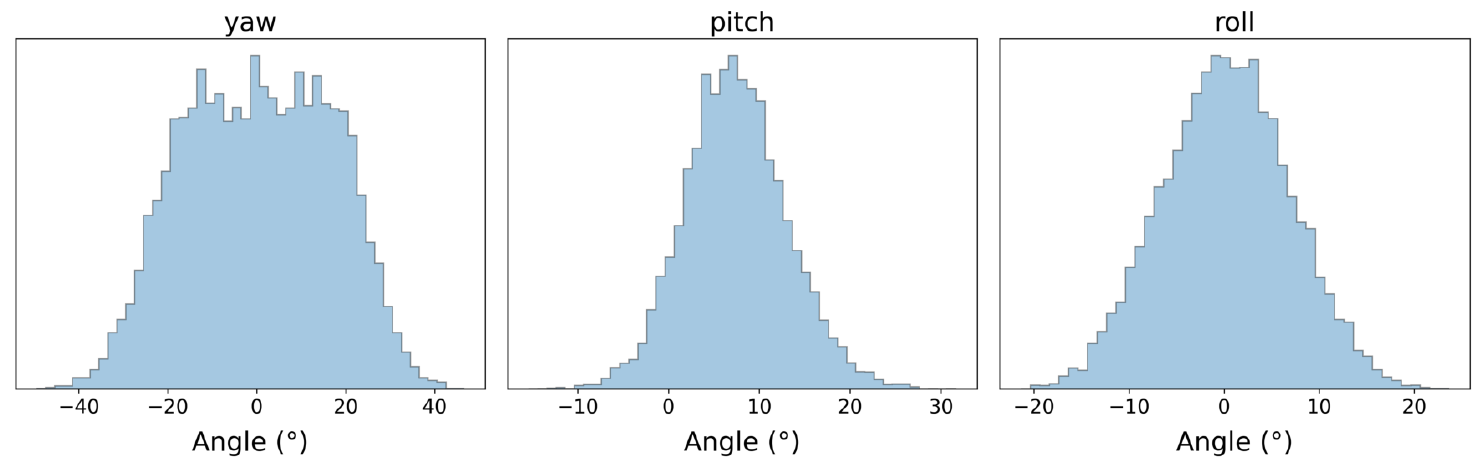}
\end{center}
  \caption{Distribution of the three Euler angles of head pose (yaw, pitch, roll) extracted using 10$\mathrm{K}$ synthetic images from our pretrained GAN on VoxCeleb1 dataset.}  
\label{fig:pose_dist}
\end{figure*}

\begin{figure}[!h]
\begin{center}
\includegraphics[width=0.5\textwidth]{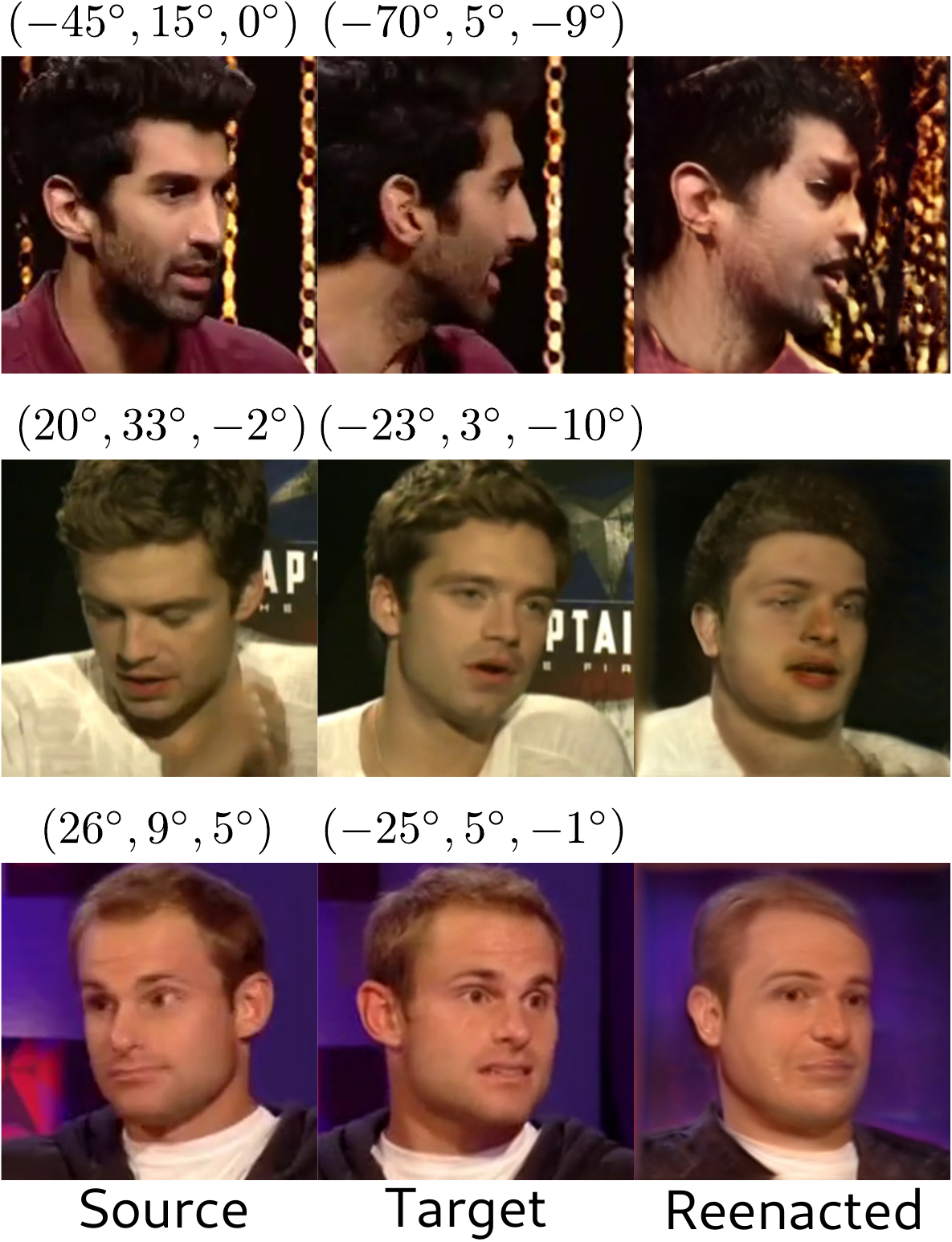}
\end{center}
  \caption{Cases where face reenactment fails, with the generated images being too blurry or contaminated with artifacts. The first two columns show the source and target images, while the reenacted images are presented on the last column.} 
\label{fig:fail_reenactment}
\end{figure}

Our method introduces a new approach to neural face reenactment, which using a 3D shape model learns the disentangled directions of facial pose in the latent space of a pretrained GAN. Our framework is simple and effective, however there are few limitations. We observed that most of failures happen when the source or target faces are on extreme head poses. Specifically, as mentioned in Section 3.1, we estimated the distribution of each element of the pose parameters $\mathbf{p}$ by randomly generating $10\mathrm{K}$ images using our pretrained generator on VoxCeleb1~\cite{Nagrani17} dataset. As shown in Fig.~\ref{fig:pose_dist}, each attribute of the head pose (yaw, pitch, roll) has a specific range. As a result, when the head pose of a real image is outside that distribution, our model often produces visual artifacts. For instance, in the first row of Fig.~\ref{fig:fail_reenactment}, the target yaw angle is $-70^{\circ}$ while on the second row the source pitch angle is $33^{\circ}$. Additionally, we notice that in some cases when the source and target head poses have large distance between them (third row in Fig.~\ref{fig:fail_reenactment}), while we are able to successfully transfer the head pose and expression, the reenacted images have visual artifacts that affect the preservation of the source identity. We attribute this to the GAN inversion process, which renders the inverted latent codes in extreme head poses less editable.

\subsection{Image editing}\label{ssec:image_editing}

\begin{figure}[h]
\begin{center}
{\includegraphics[width=1.0\textwidth]{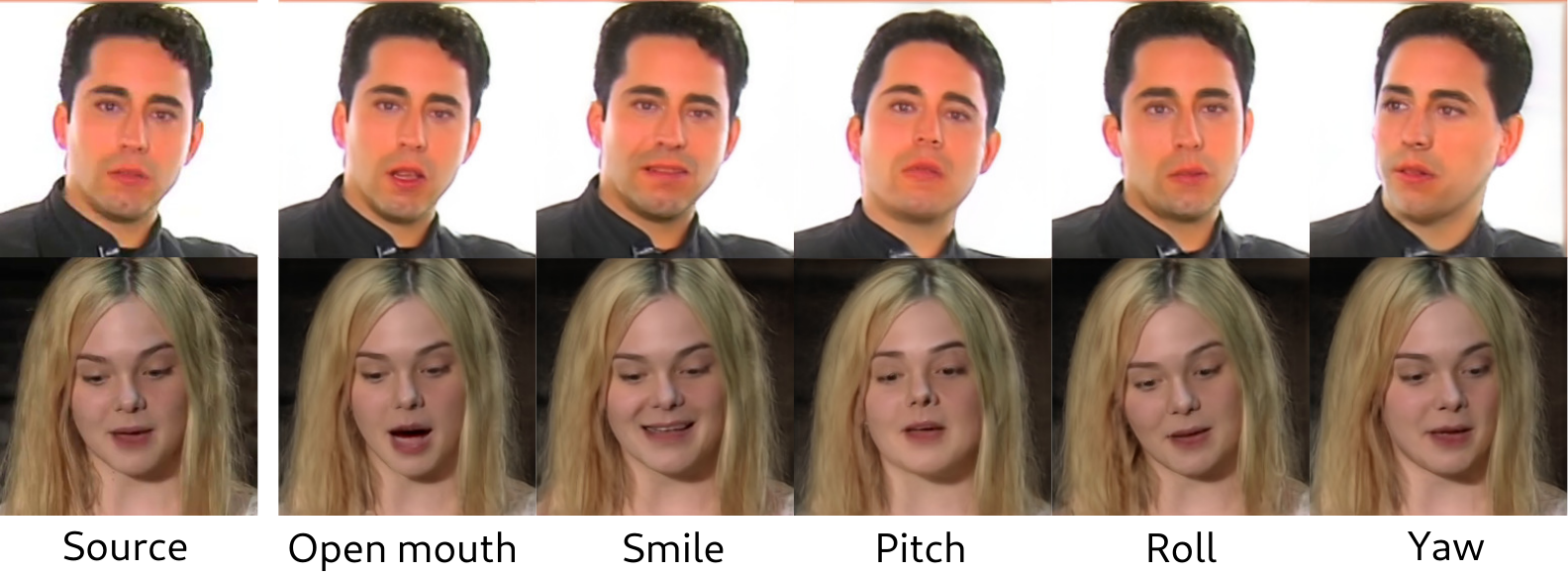}}
  \caption{Our method can perform pose and expression editing on real images. Specifically, we are able to edit an attribute by keeping all other attributes unchanged. The first column shows the source images, while the rest columns show editings of different expressions and head poses.} 
\label{fig:editing}
\end{center}
\end{figure}

\begin{figure}[h]
\begin{center}
{\includegraphics[width=0.8\textwidth]{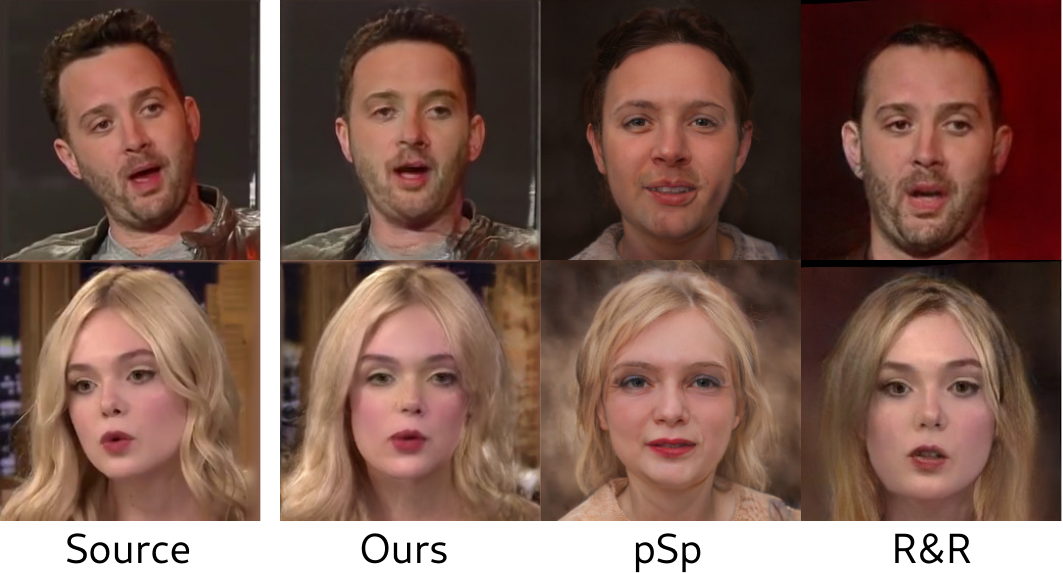}}
\end{center}
  \caption{Face frontalization examples. We perform comparisons with pSp~\cite{richardson2021encoding} and R\&R~\cite{zhou2020rotate} and we show that our method successfully perform face frontalization by preserving the identity of the source face.}\label{fig:frontal_fig}
\end{figure}

Our method is able to discover the disentangled directions of pose and expression in the latent space of StyleGAN2. Consequently, except from face reenactment, our model can perform pose and expression editing by simply setting the desired head pose or expression. Fig.~\ref{fig:editing} illustrates results of per attribute editing. As shown, our model can alter the head pose (e.g. yaw, pitch, roll) or the expression (e.g. open mouth, smile) by maintaining all other attributes unchanged. Similarly, our method can be used in the frontalization task. We compare our model with the methods of pSp~\cite{richardson2021encoding} and R\&R~\cite{zhou2020rotate} and we report both qualitative (Fig.~\ref{fig:frontal_fig}) and quantitative (Table~\ref{table:frontal}) results. Specifically, we randomly select 250 frames of different identities from the VoxCeleb test set and we perform frontalization. In Table~\ref{table:frontal}, we evaluate the identity preservation (CSIM) and the pose error between the source and the frontalized images.

\begin{table}[h]
\begin{center}
\begin{tabular}{|c||c|c|}
    \hline
    Method & CSIM $\uparrow$ & Pose $\downarrow$ \\
    \hline
    pSp~\cite{richardson2021encoding} & 0.40 & 3.0 \\
    R\&R~\cite{zhou2020rotate} & 0.45 & 3.5\\
    Ours & \textbf{0.60} & \textbf{1.2}\\
    \hline
\end{tabular}
\end{center}
\caption{Quantitative results on frontalization task. We compare our method with pSp~\cite{richardson2021encoding} and R\&R~\cite{zhou2020rotate} by evaluating the identity preservation (CSIM) and the Pose error between the source and the frontalized images.}\label{table:frontal}
\end{table}

\subsection{Additional results}\label{ssec:results}

\begin{figure*}[!ht]
\begin{center}
{\includegraphics[width=1.0\textwidth]{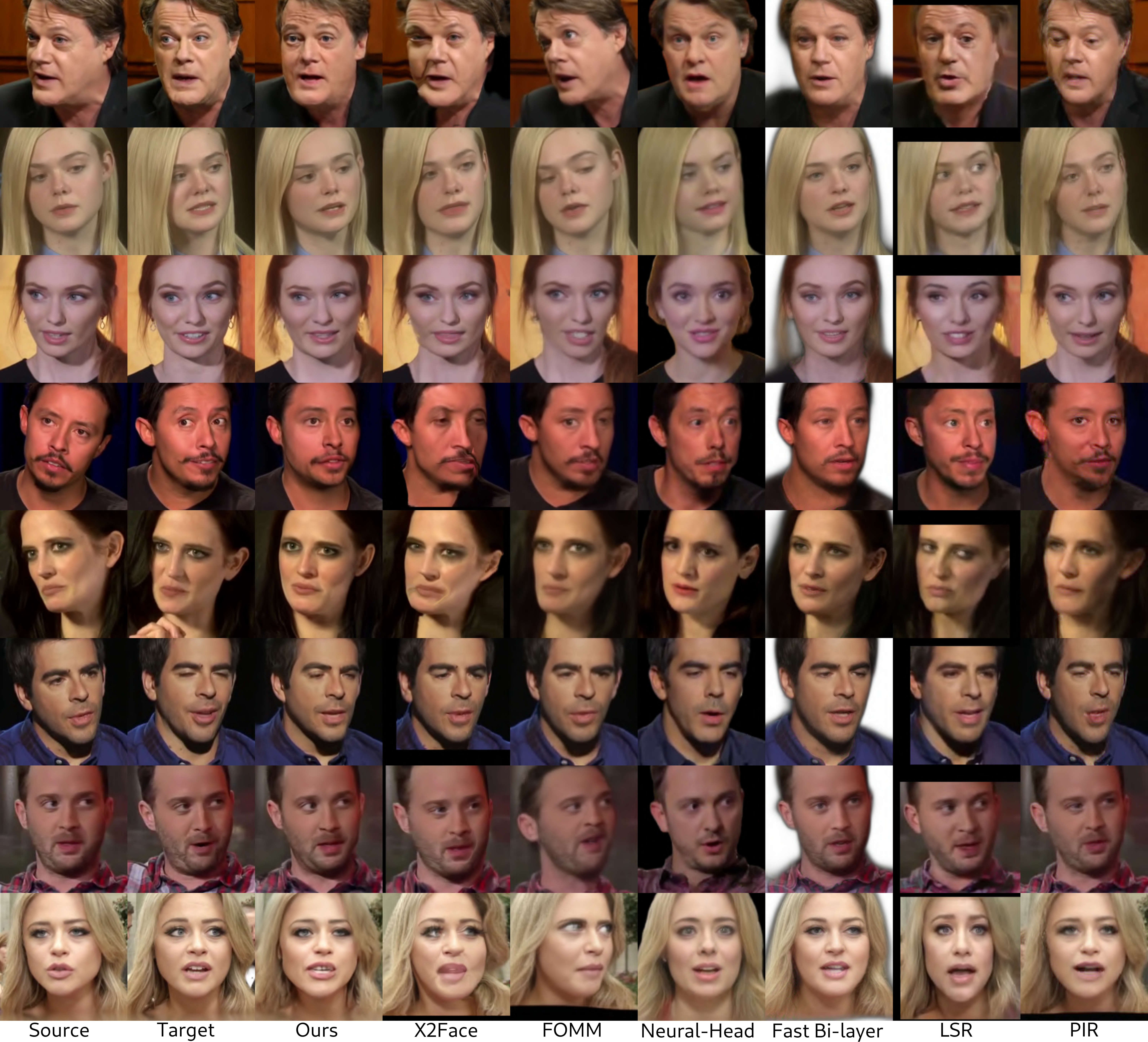}}
\end{center}
  \caption{Qualitative results and comparisons for self-reenactment on VoxCeleb1~\cite{Nagrani17} dataset. The first and second columns show the source and target faces. We compare our method with X2Face~\cite{wiles2018x2face}, FOMM~\cite{siarohin2019first}, Neural-Head~\cite{burkov2020neural}, Fast Bi-layer~\cite{zakharov2020fast}, LSR~\cite{meshry2021learned} and PIR~\cite{ren2021pirenderer}.}
\label{fig:self_reenactment_0}
\end{figure*}

\begin{figure*}[!ht]
\begin{center}
{\includegraphics[width=1.0\textwidth]{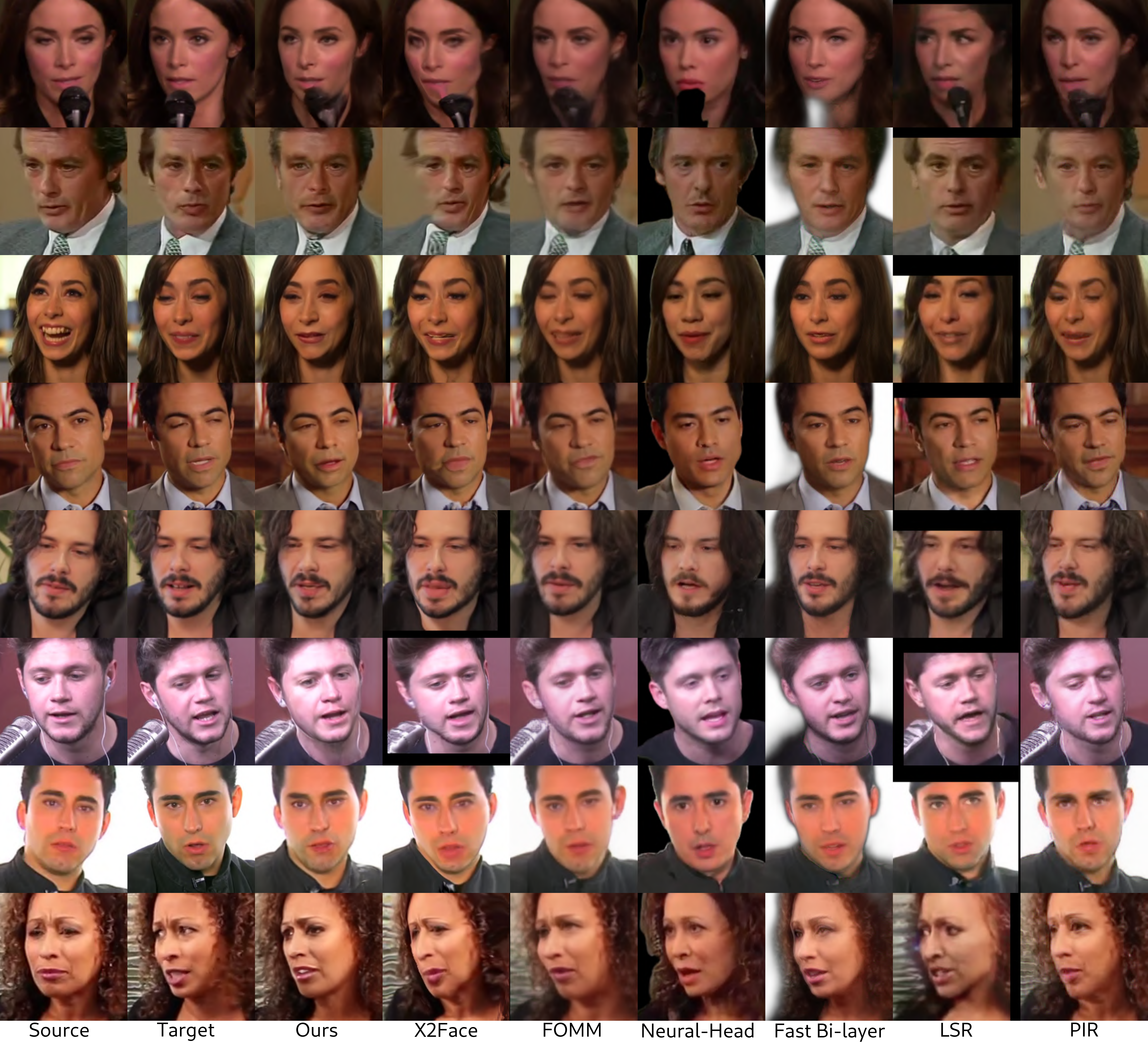}}
\end{center}
  \caption{Qualitative results and comparisons for self-reenactment on VoxCeleb1~\cite{Nagrani17} dataset. The first and second columns show the source and target faces. We compare our method with X2Face~\cite{wiles2018x2face}, FOMM~\cite{siarohin2019first}, Neural-Head~\cite{burkov2020neural}, Fast Bi-layer~\cite{zakharov2020fast}, LSR~\cite{meshry2021learned} and PIR~\cite{ren2021pirenderer}.}
\label{fig:self_reenactment_1}
\end{figure*}

\begin{figure*}[!ht]
\begin{center}
{\includegraphics[width=1.0\textwidth]{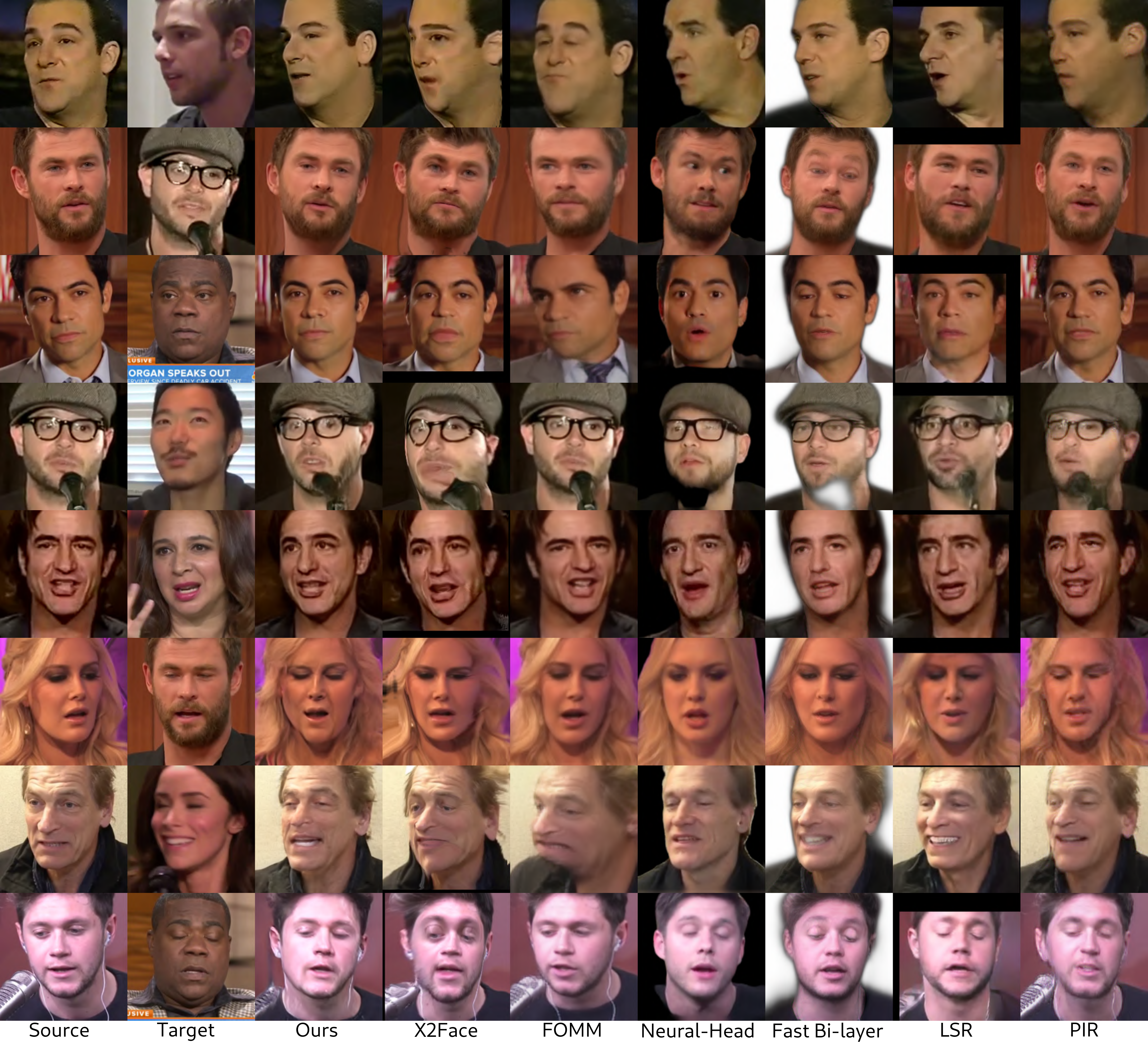}}
 \end{center}
  \caption{Qualitative results and comparisons for cross-subject reenactment on VoxCeleb1~\cite{Nagrani17} dataset. The first and second columns show the source and target faces. We compare our method with X2Face~\cite{wiles2018x2face}, FOMM~\cite{siarohin2019first}, Neural-Head~\cite{burkov2020neural}, Fast Bi-layer~\cite{zakharov2020fast}, LSR~\cite{meshry2021learned} and PIR~\cite{ren2021pirenderer}.}
 \label{fig:cross_reenactment_1}
 \end{figure*}

\begin{figure*}[!ht]
\begin{center}
{\includegraphics[width=\textwidth,height=0.9\textheight,keepaspectratio]{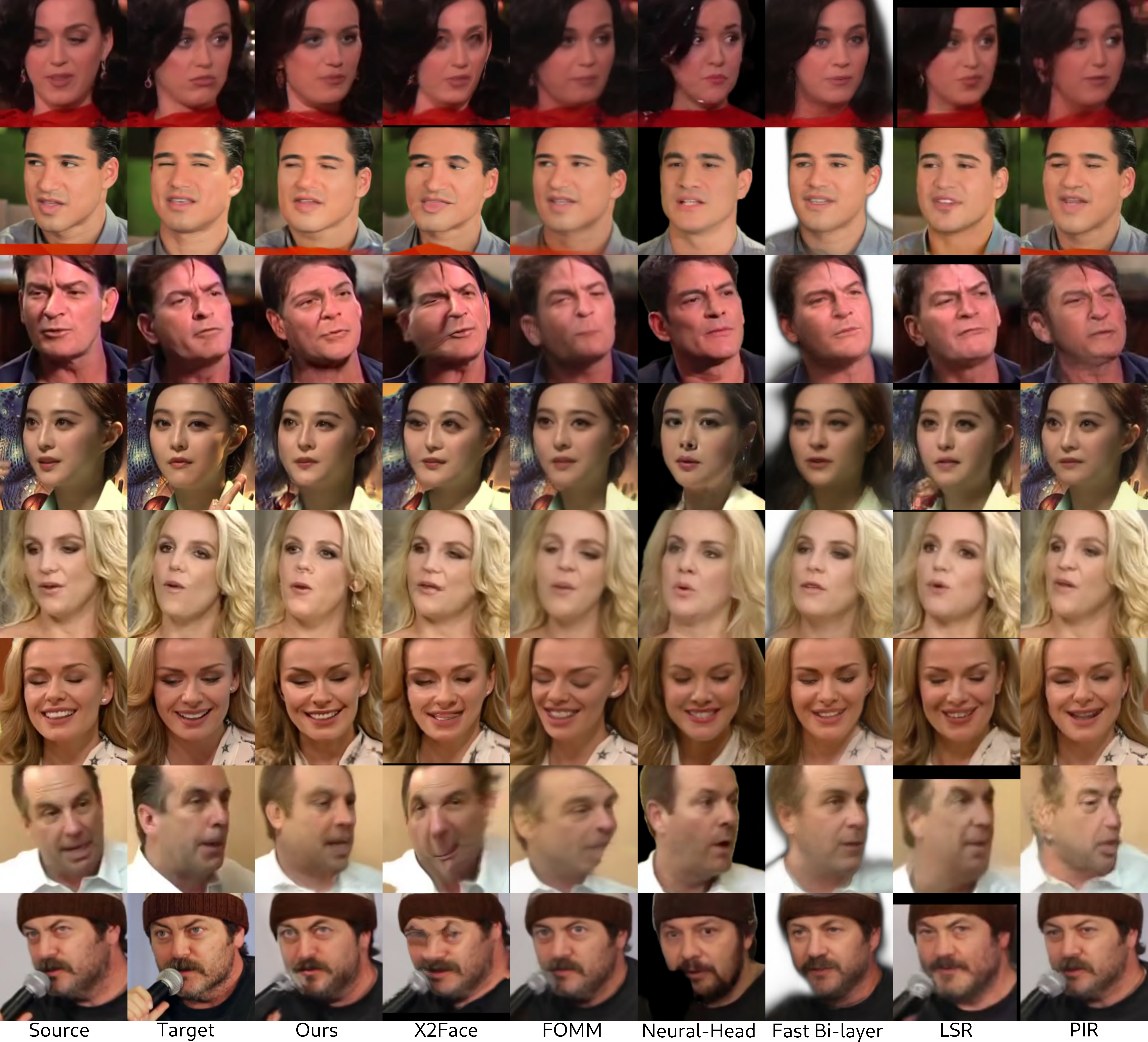}}
\end{center}
  \caption{Qualitative results and comparisons for self-reenactment on VoxCeleb2~\cite{Chung18b} dataset. The first and second columns show the source and target faces. We compare our method with X2Face~\cite{wiles2018x2face}, FOMM~\cite{siarohin2019first}, Neural-Head~\cite{burkov2020neural}, Fast Bi-layer~\cite{zakharov2020fast}, LSR~\cite{meshry2021learned} and PIR~\cite{ren2021pirenderer}.}
\label{fig:self_reenactment_2}
\end{figure*}

\begin{figure*}[!ht]
\begin{center}
{\includegraphics[width=\textwidth,height=0.9\textheight,keepaspectratio]{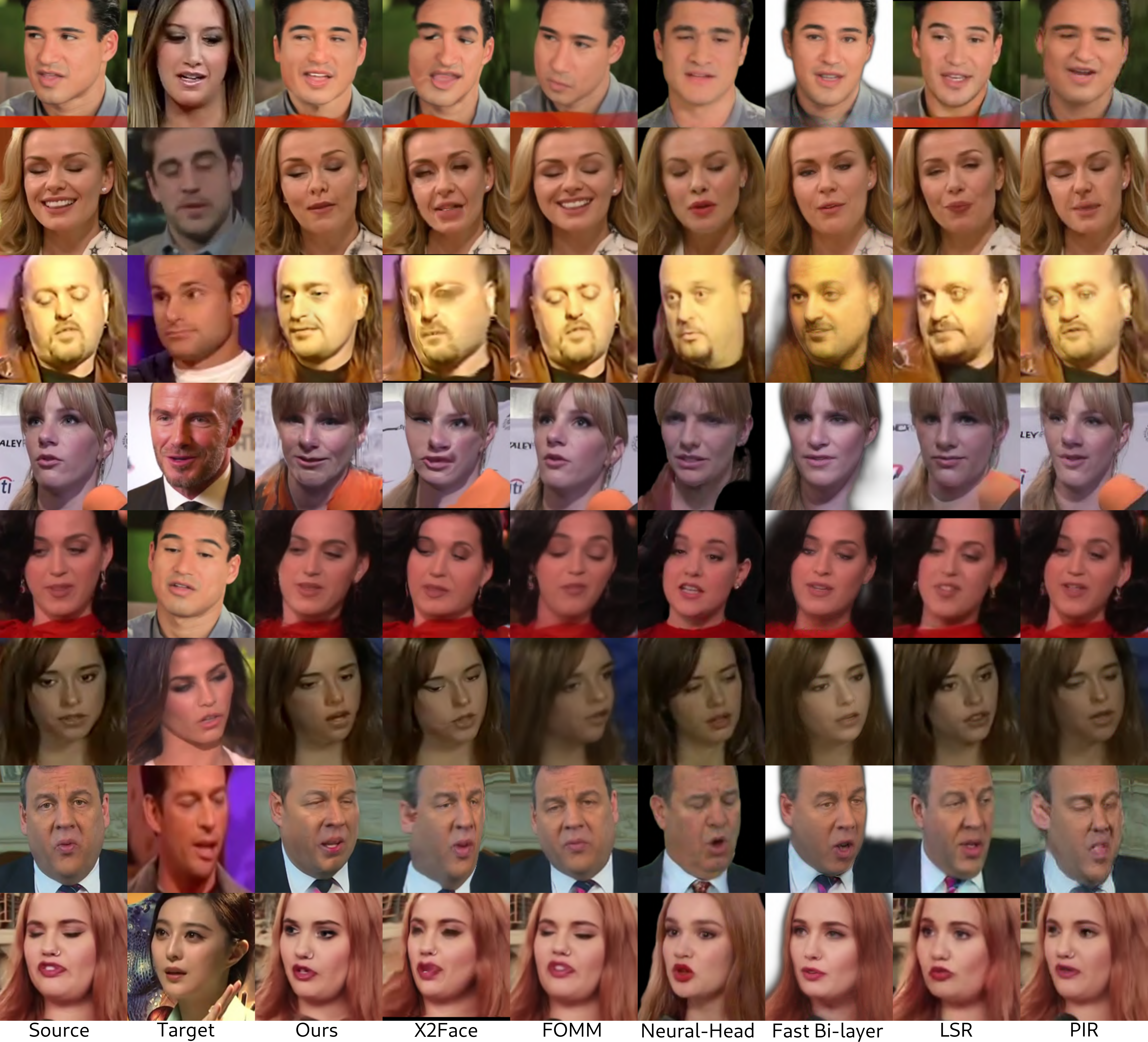}}
\end{center}
\caption{Qualitative results and comparisons for cross-subject reenactment on VoxCeleb2~\cite{Chung18b} dataset. The first and second columns show the source and target faces. We compare our method with X2Face~\cite{wiles2018x2face}, FOMM~\cite{siarohin2019first}, Neural-Head~\cite{burkov2020neural}, Fast Bi-layer~\cite{zakharov2020fast}, LSR~\cite{meshry2021learned} and PIR~\cite{ren2021pirenderer}.}
\label{fig:cross_reenactment_2}
\end{figure*}

\begin{figure*}[ht]
\begin{subfigure}{.49\textwidth}
  \centering
  \includegraphics[width=.8\linewidth]{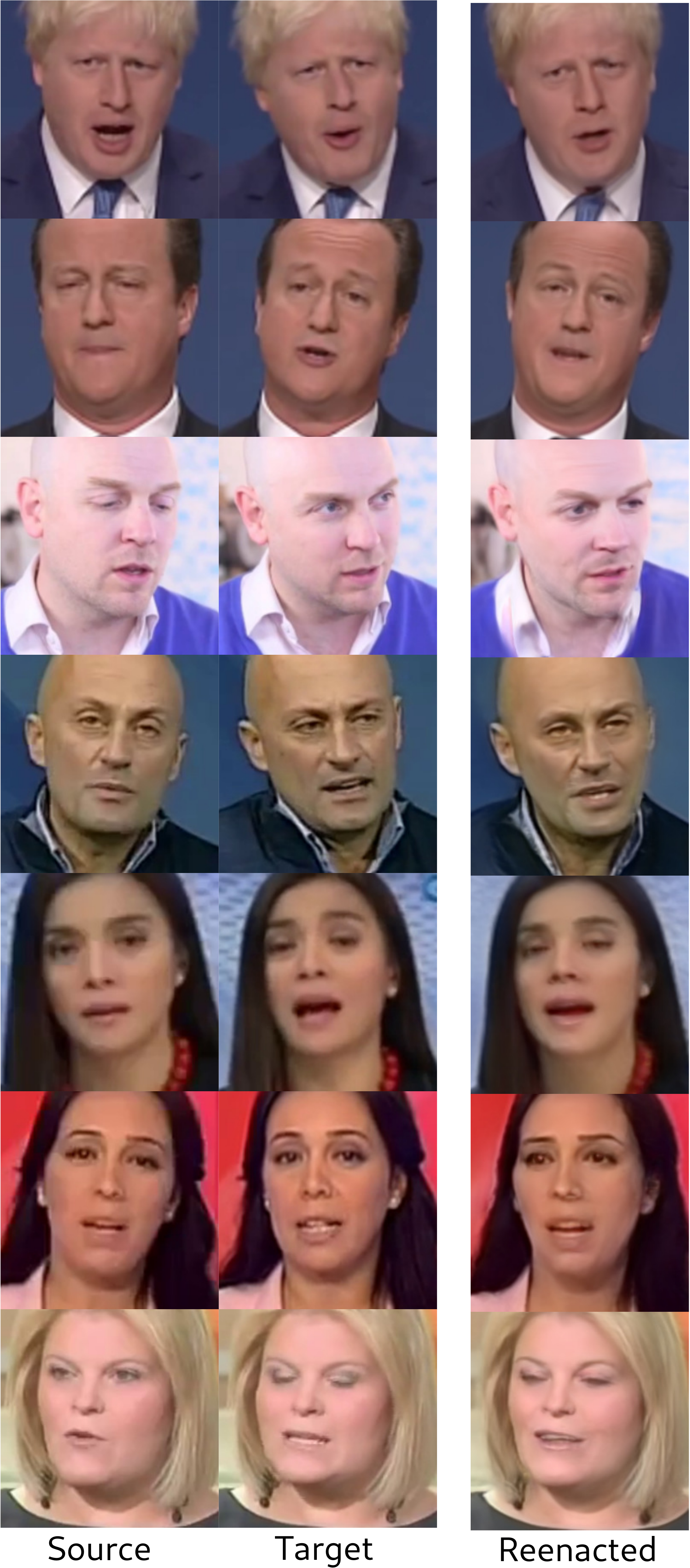}  
  \caption{Self-reenactment.}
  \label{fig:self-sub}
\end{subfigure}
\begin{subfigure}{.49\textwidth}
  \centering
  \includegraphics[width=.8\linewidth]{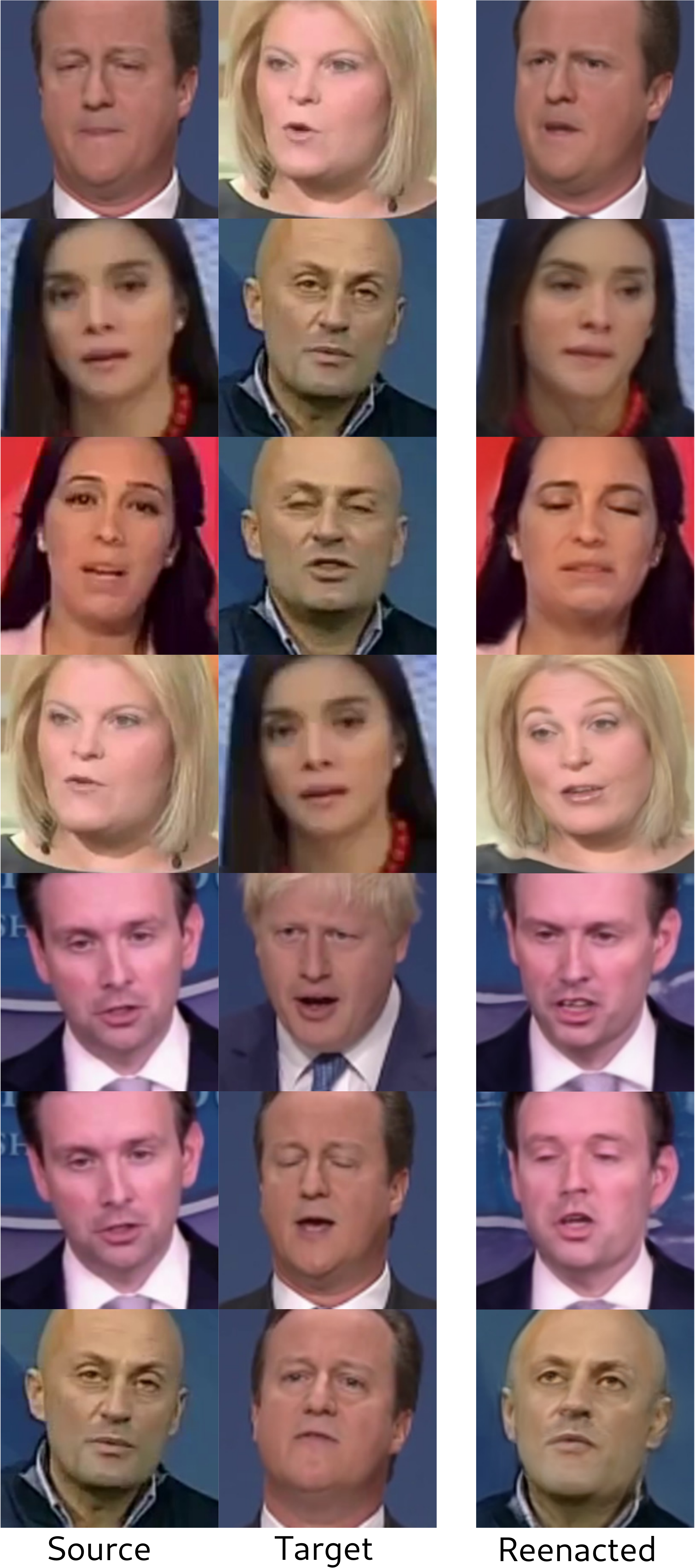}  
  \caption{Cross-subject reenactment.}
  \label{fig:cross-sub}
\end{subfigure}

\caption{Qualitative results of our method for self (a) and cross-subject (b) reenactment on FaceForensics~\cite{roessler2018faceforensics} and 300-VW~\cite{shen2015first} datasets.}
\label{fig:faceforensics}
\end{figure*}

We provide more results in self- (Fig.~\ref{fig:self_reenactment_0},~\ref{fig:self_reenactment_1}) and cross-subject (Fig.~\ref{fig:cross_reenactment_1}) reenactment in VoxCeleb1~\cite{Nagrani17} dataset and we compare our method with  X2Face~\cite{wiles2018x2face}, FOMM~\cite{siarohin2019first}, Fast bi-layer~\cite{zakharov2020fast}, Neural-Head~\cite{burkov2020neural}, LSR~\cite{meshry2021learned} and PIR~\cite{ren2021pirenderer}. Moreover, in Fig.~\ref{fig:self_reenactment_2},~\ref{fig:cross_reenactment_2} we show additional comparisons in VoxCeleb2~\cite{Chung18b} dataset. Finally, we provide \textbf{ALL} self-reenactment videos from the small test sets of VoxCeleb1 and VoxCeleb2, as defined in~\cite{zakharov2019few}, with the exception of some videos, which are not available for download\footnote{We were not able to download: id10178, id10269, id10595, id10675, id10902, id10919, id10966, id11207 from VoxCeleb1 and id00061, id00154, id01224, id03127, id04295, id04570, id04862, id04950, id05999, id08696 from VoxCeleb2.}, and 10 random pairs from each dataset for cross-subject reenactment. Furthermore, we show that our method is able to generalise well on other facial video datasets. In Fig.~\ref{fig:faceforensics} we provide results on FaceForensics~\cite{roessler2018faceforensics} and 300-VW~\cite{shen2015first} datasets both on self (Fig.~\ref{fig:self-sub}) and on cross-subject (Fig.~\ref{fig:cross-sub}) reenactment.

\begin{figure}
\begin{center}
{\includegraphics[width=0.8\textwidth]{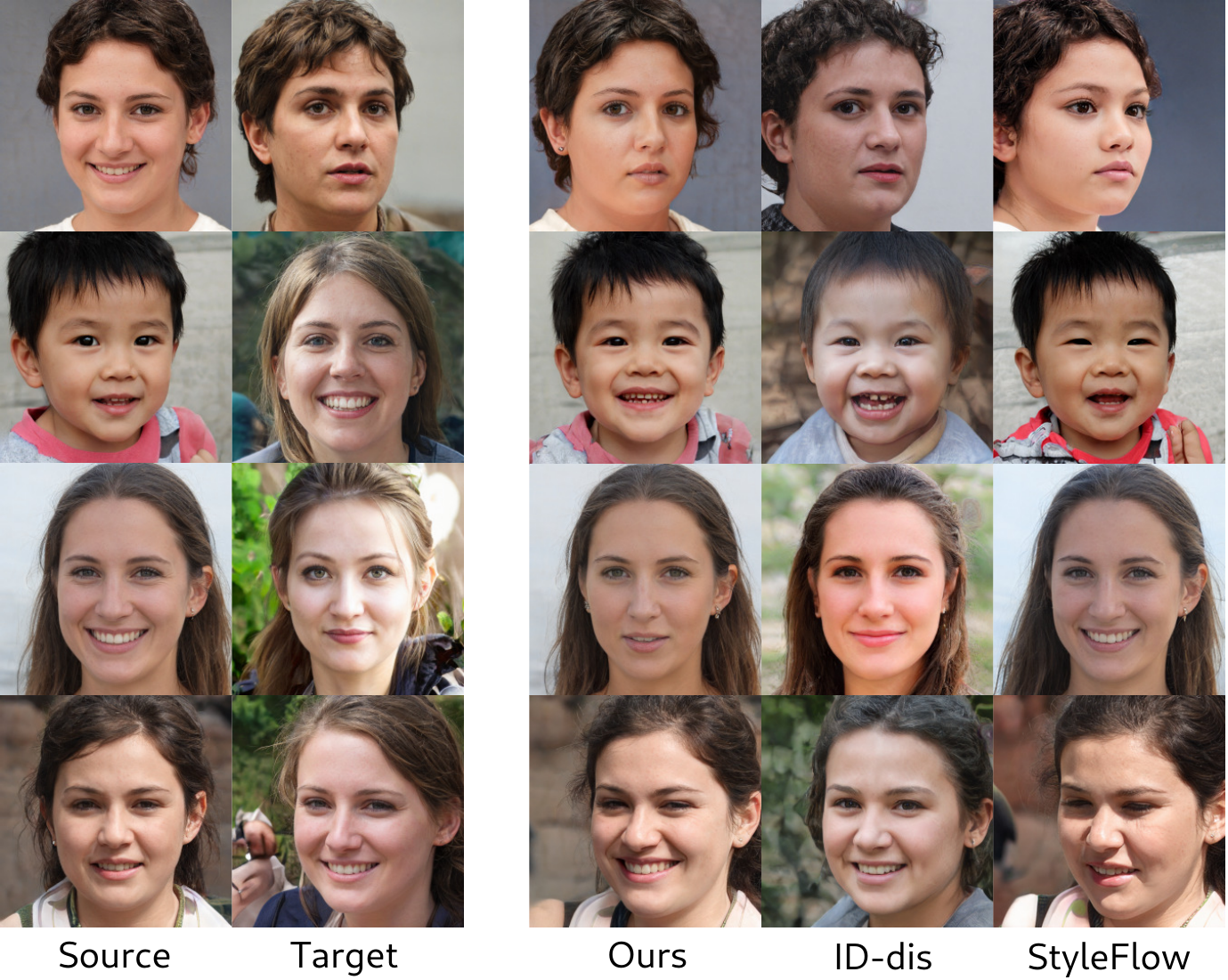}}
\end{center}
  \caption{Qualitative comparisons against ID-disentanglement (ID-dis)~\cite{nitzan2020face} and StyleFlow~\cite{abdal2021styleflow} using random source-target pairs from the small test set provided by the authors of StyleFlow~\cite{abdal2021styleflow}.}\label{fig:comparisons_synthetic}
\end{figure}

Finally, to show the superiority of our method against methods for synthetic image editing, we compare against two state-of-the-art methods, namely ID-disentanglement~\cite{nitzan2020face} and StyleFlow~\cite{abdal2021styleflow}. The authors of ID-disentanglement~\cite{nitzan2020face} introduce a method that learns to disentangle the facial pose and the identity characteristics using a pretrained StyleGAN2 on FFHQ dataset. Additionally, StyleFlow~\cite{abdal2021styleflow} is a state-of-the-art method that finds meaningful non-linear directions in the latent space of StyleGAN2 using supervision from multiple attribute classifiers and regressors. Both ID-disentanglement~\cite{nitzan2020face} and StyleFlow~\cite{abdal2021styleflow} provide pretrained models using the StyleGAN2 generator trained on FFHQ dataset~\cite{karras2019style}. Consequently, in order to fairly compare against these methods, we train our model using synthetically generated images from StyleGAN2 generator trained on FFHQ. We compare against ID-disentanglement~\cite{nitzan2020face} and StyleFlow~\cite{abdal2021styleflow} on cross-subject reenactment using synthetic images. Specifically, we use the small test set (1000 images) provided by the authors of StyleFlow~\cite{abdal2021styleflow} and we randomly select 500 image pairs (source and target faces) to perform face reenactment. In Table~\ref{table:comparisons} and in Fig.~\ref{fig:comparisons_synthetic}, we show quantitative and qualitative results of our method against ID-disentanglement~\cite{nitzan2020face} and StyleFlow~\cite{abdal2021styleflow}. As shown in Table~\ref{table:comparisons} our method outperforms all other method both on identity preservation (CSIM) and on pose transfer metrics, namely Pose, Exp. and NME. Additionally, as illustrated in Fig.~\ref{fig:comparisons_synthetic}, our method can successfully edit the source image given the target facial pose, without altering the source identity. On the contrary, ID-disentanglement (ID-dis) method~\cite{nitzan2020face} is not able to preserve the source identity, while StyleFlow fails to faithfully transfer the target head pose and expression. 

\begin{table}
\begin{center}
\begin{tabular}{|c||c|c|c|c|}
    \hline
    Method & CSIM  & Pose  & Exp. & NME\\
    \hline
    ID-disentanglement~\cite{nitzan2020face}  & 0.56 & 2.0 & 0.12 & 12.0\\
    StyleFlow~\cite{abdal2021styleflow} & 0.67 & 2.6 & 0.13 & 16.0 \\
    Ours  & \textbf{0.80} & \textbf{1.1} & \textbf{0.09} & \textbf{10.1} \\
    \hline
\end{tabular}
\end{center}
\caption{Quantitative comparisons against two state-of-the-art methods for synthetic image editing, namely ID-disentanglement~\cite{nitzan2020face} and StyleFlow~\cite{abdal2021styleflow}. For CSIM metric, higher is better ($\uparrow$), while in all other metrics lower is better ($\downarrow$).}\label{table:comparisons}
\end{table}

\end{document}